# Relevant Knowledge First

## Reinforcement Learning and Forgetting in Knowledge Based Configuration

2001-09-19

Ingo Kreuz, DaimlerChrysler AG, Stuttgart, Germany

Dieter Roller, University of Stuttgart, Germany

## Abstract

In order to solve complex configuration tasks in technical domains, various knowledge based methods have been developed. However their applicability is often unsuccessful due to their low efficiency. One of the reasons for this is that (parts of the) problems have to be solved again and again, instead of being "learnt" from preceding processes. However, learning processes bring with them the problem of conservatism, for in technical domains innovation is a deciding factor in competition. On the other hand a certain amount of conservatism is often desired since uncontrolled innovation as a rule is also detrimental.

This paper proposes the heuristic RKF (Relevant Knowledge First) for making decisions in configuration processes based on the so-called *relevance* of objects in a knowledge base. The underlying relevance-function has two components, one based on reinforcement learning and the other based on forgetting (fading). Relevance of an object increases with its successful use and decreases with age when it is not used. RKF has been developed to speed up the configuration process and to improve the quality of the solutions relative to the reward value that is given by users.

## 1 Motivation

Before we begin, we will give a short introduction to *knowledge based configuration*, the field for which *Relevant Knowledge First* (RKF) has been developed, followed by possible improvements and effects from cognitive psychology which will lead us into the introduction of RKF.

## 1.1 Knowledge-Based Configuration

Several definitions of configuration have been proposed (e.g. [McDermott 1982], [Mittal 1989] ). It's interesting that sometimes configuration is more generally called "design" (e.g. [Mittal et al. 1986], [Tong et al. 1992], [Roller et al. 1998]). A definition that goes with our understanding of configuration in this article has been presented by Günter. In [Günter 1995] he defines configuration as *a task of synthesis of domain objects* and in [Günter et al. 1999a] the characteristics of configuration are described as followed: *„Configuration tasks have the following characteristics: A set of objects in the application domain and their properties (parameters), a set of relations between the domain objects where the taxonomical and compositional relations are of particular importance for configuration, a task specification (configuration objectives) that specifies the demands a created configuration has to accomplish and control knowledge about the configuration process."*

Today's commercial configuration tools (see [Günter et al. 1999b]) are restricted mostly to checking whether a configuration is technically feasible. As a rule not only the selection of components must be made manually but also a conflict resolution. Several known methods and in particular heuristics exist however with which it could be possible to configure to a greater extend without user interaction. Some of these are described in [Cunis et al. 1991] and [Günter 1995]. Their feasibility has been proved in development prototypes and will be used partly in future commercial tools (see summary in [Günter et al. 1999b]).

The configuration or domain knowledge in configuration systems is usually structured in one of the following ways (see [Günter et al. 1999a]): *Object or frame-based approach* (e.g. in KONWERK [Günter 1995], SCE [Haag 1998]), *resource oriented approach* (e.g. in COSMOS [Heinrich et al. 1996]), *rule oriented approach* (e.g. in XCON [McDermott 1982]) and *mixed forms*.

In this article we will introduce our heuristic RKF in context of the first approach (object/frame based) because our application uses this form of representation. However RKF can always be used to make decisions, regardless of the form of representation. Whenever we speak of objects in this article it can be generalized to any piece of information or knowledge. In section 3 we give an overview of the types of objects we intend to use RKF.

## 1.2 Possible Improvements for Knowledge Based Configuration

The performance and quality of knowledge-based methods are essentially determined by the search methods used, e.g. the methods to draw back decisions (revision, e.g. backtracking) and the heuristics employed. In the following, characteristics of present knowledge-based methods will be considered which offer the potential for improvement. This list, which is probably incomplete, gave us a starting point for developing RKF.

- *Making "correct" decisions:* All heuristics aim to make the best possible decisions. In other words, revision should be avoided in order to accelerate the knowledge-based process. At the same time the best possible solutions concerning one or several optimization criterions are to be found without investigating the whole search space.
- *Learning from successful preceding processes:* Feedback from a solution on the knowledge base is desirable: The knowledge-based system similarly "learns" the way human experts do from preceding solutions. This "experience" can be applied to new solutions.
- *Avoid conservatism in spite of "learning":* One danger of self-learning methods is conservatism. If the main reason for making a decision is because "it has always been done in this way", then a system is not in a position to take new factors adequately into consideration. Innovation is an important economic factor especially in technical areas.
- *Versions control of knowledge bases:* Version control is not typical for objects in knowledge bases. Obviously they can be modeled in general systems for knowledge representation[1]. This modeling is mostly nontrivial if old versions of knowledge should be left in the knowledge base. For example if old versions are already in use in the original system which has to be *re*configured, they must also be accessible. In general it is often better to give preference to new knowledge, without deleting the old.
- *Maintenance of the Knowledge Base:* Knowledge bases can become extremely large after a period of time. The larger the knowledge base, the larger the search space and the slower the average running time of a knowledge-based search algorithm. The need for knowledge base maintenance is also mentioned in chapter 4.4 of [Bense et al. 1995]. It should be pointed out that on principle it is risky, to completely erase knowledge, simply because it has not been needed for some time.

## 1.3 Effects from Cognitive Psychology

Although incomplete, the list of possible improvements in section 1.1 leads us on to the question of how humans handle knowledge. Still an enormous number of NP-complete problems are "solved" every second by humans who are able to fall back on an enormous knowledge base. In this context David Waltz (see [Waltz 1999]) sees the human brain as an important means of survival e.g. to work out the answer to significant questions such as *"can I eat it or can it eat me?"*. He comes to the conclusion that the "importance" of information is crucial to how the world model changes and whether information should generally be stored.

[1] in chapter 2 of [Karbach et al. 1990] a definition of "knowledge representation systems" can be found

In his book [Anderson 2000] Anderson describes a model where the human brain is able to handle enormous amounts of information and therefore large search spaces. From this model we shall firstly consider the so-called ***activation** of a memory trace*. It indicates how accessible information is for a current problem, i.e. how fast and with what probability information can be accessed. Anderson puts forward analyses which show that along with the accessed information, related information is also activated. Every time we use a memory trace its activation increases a little. In this model the gradual effect of training is recorded for each piece of information through a so-called *action potential*. This leads to the effect that during problem solving information is *in principle varyingly good*, that's to say *varyingly quickly* accessible. If a piece of information is needed frequently, its *action potential* increases. We will call this "training" or "learning". The relationship between the quantity of exercise and the access efficiency (e.g. measured as reaction time) results in a power function which is known in cognitive psychology terms as the *power law of learning*. Figure 1 has been taken from [Anderson 2000] and shows some results from J. M. Blackburn. In 1936 he investigated the effect of training when subjects added numbers in 10,000 exercises.

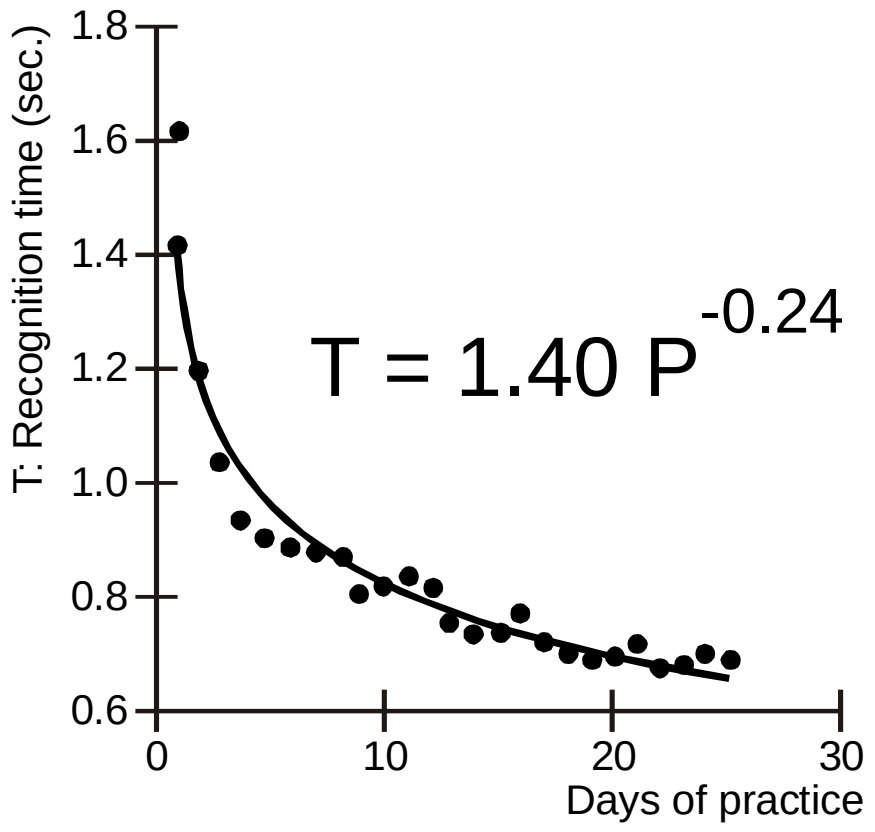

*Figure 1: Blackburn's results concerning the* ***power law of learning***

Learning information counteracts forgetting: If information is not used over a period of time, its action potential fades. Experiments show that forgetting can also be described as a power function, which amongst other things could be explained by the decaying processes of the neural connections.

Anderson refers to the experiments carried out by W.A. Wickelgren, who asked subjects to learn lists of words. He investigated the probability of the words being recalled after various time spans (one minute to 14 days). This probability served as a measure for the memory's performance *d'* and is represented in Figure 2 as a function of the time taken to recall the list of words.

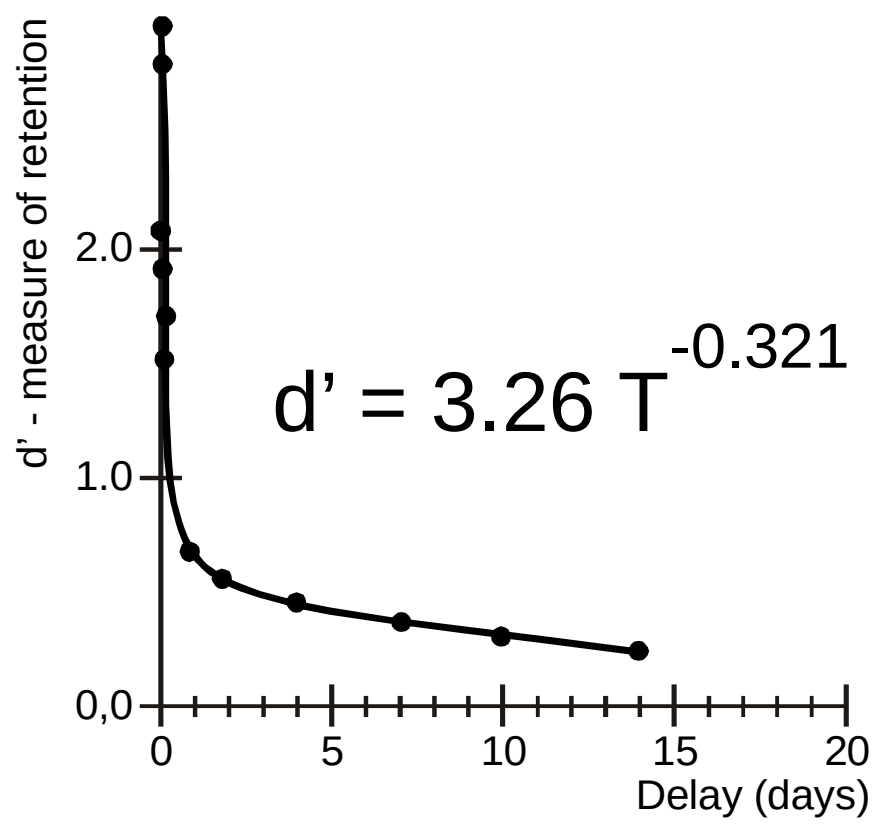

*Figure 2: Wickelgren's results concerning the* ***power law of forgetting***

The effects described show that people can concentrate on relevant information (high activation level of a certain memory trace) to solve concrete problems and they make relevant information more or less accessible through the processes of "learning" and "forgetting". The assessment of the relevance of information could be a deciding factor in managing the vast amount of information stored in the human brain.

This idea encouraged us to develop a heuristic in the area of knowledge based configuration, which can identify relevant information for a given problem and thereby drastically accelerate the solution process. In order to find a solution not only quickly, but also of "good" quality, the principle suggested by Anderson had to be adapted.

# 2 Relevant Knowledge First (RKF)

## *2.1 Principle of RKF*

For the assessment of the relevance of objects there are two deciding factors which correspond to the antagonism between conservatism and innovation: On the one hand objects are very likely to be *relevant* again if they have already been useful for similar tasks (see *training* in section 2.4.1) and objects that did not help to find solutions will probably not help in the future (see *forgetting* in section 2.4.2). On the other hand *new* objects should be taken into consideration in order to avoid conservatism: Because objects can be forgotten it is possible that new objects that "prove their worth" can replace objects that have been most relevant so far. As an assessment for objects we calculate their so-called *relevance* from the time since they were last used (during forgetting phases) and the rewards obtained by users (during learning phases).

With RKF the search for solutions is supported by *preferring relevant* objects: Every time a decision has to be made, the relevance of all objects *in question* $o \in \boldsymbol{O}$ at the actual time $t$ is calculated and one object is selected by a random generator, where the probability to choose an object $o$ is proportional to its relevance.

$$P(o, t, c) := \frac{\mathrm{rel}(o, t, c)^{v}}{\sum_{i=1}^{|\boldsymbol{O}|} \mathrm{rel}(o_i, t, c)^{v}} \quad o, o_i \in \boldsymbol{O} \qquad [1]$$

| | |
|---|---|
| $o$ | an object "in question" |
| $\boldsymbol{O}$ | set of all objects in question |
| $t$ | actual time |
| $c$ | context ("task class") for relevance (see 2.3) |
| $\mathrm{rel}(o_j, t, c)$ | the calculated relevance of $o$ at time $t$ in context $c$ (see 2.4) |
| $v$ | parameter to control the conservatism for selections ("variance") |

After a solution has been found, the user of the configuration system assesses the solution by giving a reward for all objects of the solution. For subsequent search processes the relevance increases for successfully used objects proportional to their reward (train) and decreases for all other objects (forget).

## 2.2 Relevance

The term "relevance" is also known in the field of information retrieval. In this context "relevance" is used as a measure of how good a list of documents found by an information retrieval system answers a user's query ( [Schamber et al 1990], [Crestani 1995], [Anderson 1998] ). Schamber marks *"relevance is currently recognized as a multidimensional and dynamic concept extending beyond the traditional definition so closely associated with topic-matching and an identification of relevance as the relation between a document and a search question"*. We also think that the relevance of knowledge is dynamic in configuration tasks in technical domains because facts change e.g. as a result of innovations. A more common definition of relevance is given by Pearl [Pearl 1988]: *"Relevance is a relationship indicating a potential change of belief due to a specified change in knowledge. Two propositions A and B are said to be relevant to each other in context C if adding B to C would change the likelihood of A"*. When we use relevance we will derive a probability value from it and we will also define a context for the relevance-values that we call *task class*. The system described by Crestani [Crestani 1995] learns relevance-values using feedback mechanisms. As described above RKF includes a learning portion that uses ideas from reinforcement learning. In the context of forgetting, Lin and Reiter give a definition of relevance [Lin et al. 1994]: *"Let T be a theory, q a query and p a ground atom. We say that p in T is irrelevant for answering q, iff T and forget(T, p) are equivalent"*, where forget(T, p) means the deletion of p from T. We have a different interpretation of forgetting (see 2.4.1), which in our context describes a continuous process that is similar to "fading" in the system "aHUGIN" [Olesen et al. 1992]; but we will also suggest a mechanism for deleting knowledge dependent upon its relevance (see "Maintenance of Knowledge Bases" in section 3).

Very similar to our understanding of relevance is *utility* of control rules (which can also be seen as objects) that is discussed in [Rich et al. 1991]: *"As we add more and more control knowledge to a system, the system is able to search more judiciously. [...] However deliberating about which step to take next in the search space, the system must consider all the control rules. If there are many control rules, simply matching them all can be very time-consuming [...] PRODIGY maintains a utility measure for each control rule. This measure takes into account the average savings provided by the rule, the frequency of its application, and the cost of matching it."* Also the *value* of information introduced by Markovitch and Scott [Markovitch et al. 1988] is similar to our relevance. In their L*F heuristic the "length of a macro and the number of times information was used in the evaluation phase" serves to select knowledge from a vast amount of automatically generated rules. In all these considerations the value of information is used to estimate to what extent it will probably help to solve tasks in the future.

We will now put forward our definition of relevance with a sectional representation: The relevance at the time of *t* of an object *o* in the context of a task class *c* is calculated as a function of time since a last access (forget, if *o* is **not** part of the solution) and the rewards given by a user (train, if *o* **is** part of the solution).

$$\mathrm{rel}(o, t, c) := \begin{cases} \mathrm{train}(o, t, c) & \text{if } o \textbf{ is} \text{ part of the solution} \\ \mathrm{forget}(o, t, c) & \text{if } o \text{ is } \textbf{not} \text{ part of the solution} \end{cases} \qquad [2]$$

The power functions in section 1.3 show desirable characteristics for train(*o, t, c*) and forget(*o, t, c*) in technical domains. They were therefore a starting point for the "train" and "forget" functions, which we will introduce in section 2.4.

## 2.3 Task Classes

As described in 2.2 we always understand the relevance of an object relative to a context *c* that we call *task class*. For task classes the following is true:

1. task classes are the contexts for the relevance of objects, i.e. every object has one relevance value per task class
2. task classes form a partitioning of the set of all possible configuration tasks of a domain, i.e. every task can be unambiguously assigned to one task class
3. the rewards given for an object as part of a solution are (nearly) constant or change slowly in the context of one single task class as long as the domain is not changed (objects or dependencies are added or removed). Changes of the domain are seldom

One target of our heuristic RKF is to learn, which objects get the highest rewards, i.e. each task class implies the task specification which will be learnt by RKF. The highest relevance corresponds to the highest rewards of an object that it has yielded recently. It is necessary to have multiple task classes (axiom 1) because conflicting goals can exist in a domain which we take into consideration by multiple relevance values. When an actual configuration task is started a task class has to be selected as the context for the task. The second axiom ensures that this selection is unambiguous. Axiom 3 ensures that there is enough time to learn the object with the highest rewards per task class. The permanent learning enables RKF to follow slow changes. The forgetting portion ensures that an abrupt change of rewards which is allowed as a consequence of domain changes can result in a new “most relevant” object. These abrupt changes should occur seldom, because RKF needs time to learn the altered rewards. What “slowly” and “seldom” means, is determined by the parameters of the relevance function (that will be introduced in section 2.4). These parameters control how fast or how conservative the learning and forgetting behavior of RKF is and how much time is needed to learn the objects with the best rewards.

The above statements lead to the problem of finding proper task classes for a domain. Interestingly, task classes already exist in most configuration domains: User or target groups of a product.

For example task classes of a personal computer domain can be {Home-PC, Server-PC, CAD-PC , ...}.

If no such groups exist for a domain, a second indicator for task classes emerges from axiom 3: The task classes can result from the combinations of various global optimization objectives. For example if the optimization objectives “price” and “performance” are decisive in a domain, the set of task classes can be generated by combining the objectives for price and performance such as “high”, “low” and “don’t care”. The result would be $3 \cdot 3 = 9$ task classes: “low price / high performance”, “don’t care the price / high performance”, “high price / high performance”, “low price / don’t care performance” etc.

It is also possible to start with a course partitioning of the set of tasks and to refine the task classes later. Therefore the task classes can be further partitioned with the following algorithm: Divide *c* into *c1* and *c2*: Rename *c* into *c1*. Create a new task class *c2*. Copy all relevance information for all objects from *c1* to *c2*. Modify the partitioning function to classify tasks from *c* from now on as *c1* or *c2* (exclusively).

The following example shows that the refinement of task classes is very natural: The above mentioned task class “Home-PC” which was probably a typical target group some years ago can nowadays be refined to “Game-PC”, “Internet-PC”, “Private-Office-PC”, “Video-PC” etc.

## 2.4 Relevance Function for Technical Domains

Equation [2] shows that training and forgetting are effects that alternate over time. For the following considerations we have used the number of configuration runs as time. This discrete time measure especially simplifies the description of learning phases because a user reward is only available in the context of a configuration run. With this method, time has to be determined separately for each task class because every configuration run is in context of exactly one task class, so the time goes on only for this task class. In this paper we will write *t* for time as a substitution for $t_c$ (time *t* in context of task class *c*) to make the equations clearer.

Initially we will consider the processes of training and forgetting separately.

### 2.4.1 Function for Training

The learning portion of RKF is very similar to reinforcement learning (see [Kaelbling et al. 1996]). This learning model uses the following standard model:

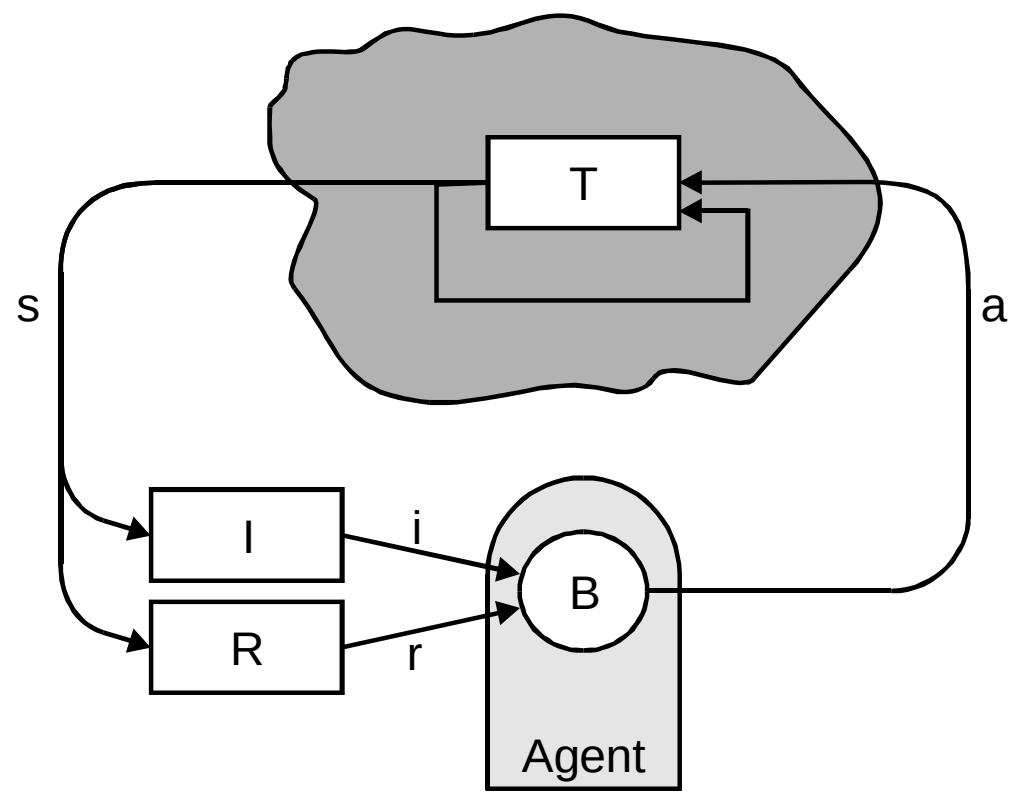

*Figure 3: The standard reinforcement-learning model*

The model consists of a discrete set of environment states ***S***, a discrete set of agent actions ***A***, a set of scalar reinforcement signals, typically {0, 1} or the real numbers, a state transition probability function T: s' = T(s, a), an input function I, which determines how the agent views the environment state I: i = I(s) (usually the identity function), a reward function R: r = R(s) and a behavior function B to choose actions B: a = B(i, r). *"The agent's behavior, B, should choose actions that tend to increase the long-run sum of values of the reinforcement signal.[...] The simplest possible reinforcement-learning problem is known as the k-armed bandit problem, which has been the subject of a great deal of study in the statistics and applied mathematics. The agent is in a room with a collection of k gambling machines (each called a "one-armed bandit" in colloquial English). The agent is permitted a fixed number of pulls, h. Any arm may be pulled on each turn. The machines do not require a deposit to play; the only cost is wasting a pull playing a suboptimal machine. When arm i is pulled, machine i pays off 1 or 0 according to some underlying probability parameter pi, where payoffs are independent events* *and the pis are unknown."* Kaelbling, Littman and Moore use the term "action" to indicate the agent's choice of arm to pull. It is very important to note that bandit problems fit the definition of a reinforcement-learning environment with a single state with only self transitions.

A decision between *all information in question* during a configuration process using RKF can be compared to the k-armed bandit problem: pulling arm i means selecting an object *o* in question. The reward or payoff is given by the user's rating *reward(o, t, c).* The use of task classes assures that the rating for an object selection remains (axiom 3 in section 2.3).

Different to the reinforcement learning model described in [Kaelbling et al. 1996] we do not linearly sum up the rewards but use an exponential function to calculate the relevance from the rewards given by users. We are motivated to do so by looking at results of Cognitive Psychology (see 1.3). The Power Function of Learning shows characteristics which also seem to be adequate for technical domains:

- an object seems to be important if it is used frequently. The "access time" should be reduced, i.e. it should preferentially be selected. In other words, its relevance should be increased
- training increases relevance starting from the last relevance value which results from the preceding forgetting phase lastForgetRel(*o, t, c*) and approaches 1 (=100%)
- The first accesses make information relevant quicker than later accesses. If for example the same object is used in 10,000 configuration processes of a particular task class, ten further accesses no longer have particular importance. Experts have thus learnt that this object is suitable in the context of the task class
- Relevance approaches a maximum value

We reduce the range of values to [0, 1]. Since we calculate the 'relevance' in contrast to the 'access time' of Figure 1, the function has to be horizontally reflected. This is based on the fact that a high relevance corresponds to a small access time.

Our exponential function for training is modeled by the following recursive formula: The difference between a last relevance value rel(*o, t*-1, *c*) and the maximum possible relevance (100%) is divided by a "basis" $b_t$. Then the last relevance value is increased by the resulting value. The user's rating reward(*o, t, c*) $\in$ [0,1] is considered by multiplying it with the increasing amount, i.e. the user's reward is the percentage value for which the increment of last relevance value is to be performed. We have used this recursive version of the train function in our implementation because it is efficient to implement and easy to understand.

***Definition 1:***

The effect of "training" for an object *o* in technical domains can be described by the following function:

$$\text{train: rel}(o, t, c) = \text{rel}(o, t\text{-}1, c) + \frac{1 - \text{rel}(o, t\text{-}1, c)}{b_t} \cdot \text{reward}(o, t, c) \text{ starting with lastForgetRel}(o, t, c) \quad [3]$$

with

| | | |
|---|---|---|
| $\text{rel}(o, t, c) \in [0,1]$ | relevance for an object *o* at time *t* in the context of task class *c* | [4] |
| $\text{reward}(o, t, c) \in [0,1]$ | the reward during the actual learning phase | [5] |
| $b_t \in (1,\infty)$ | "basis" for the train function determining the "speed" of training. The larger $b_t$ the slower is training. Typical values lie between one and three. | [6] |
| $\text{lastForgetRel}(o, t, c) \in [0, 1]$ | last relevance value of *o* before *t* when the last forgetting phase ended in the context of the task class *c* | [7] |

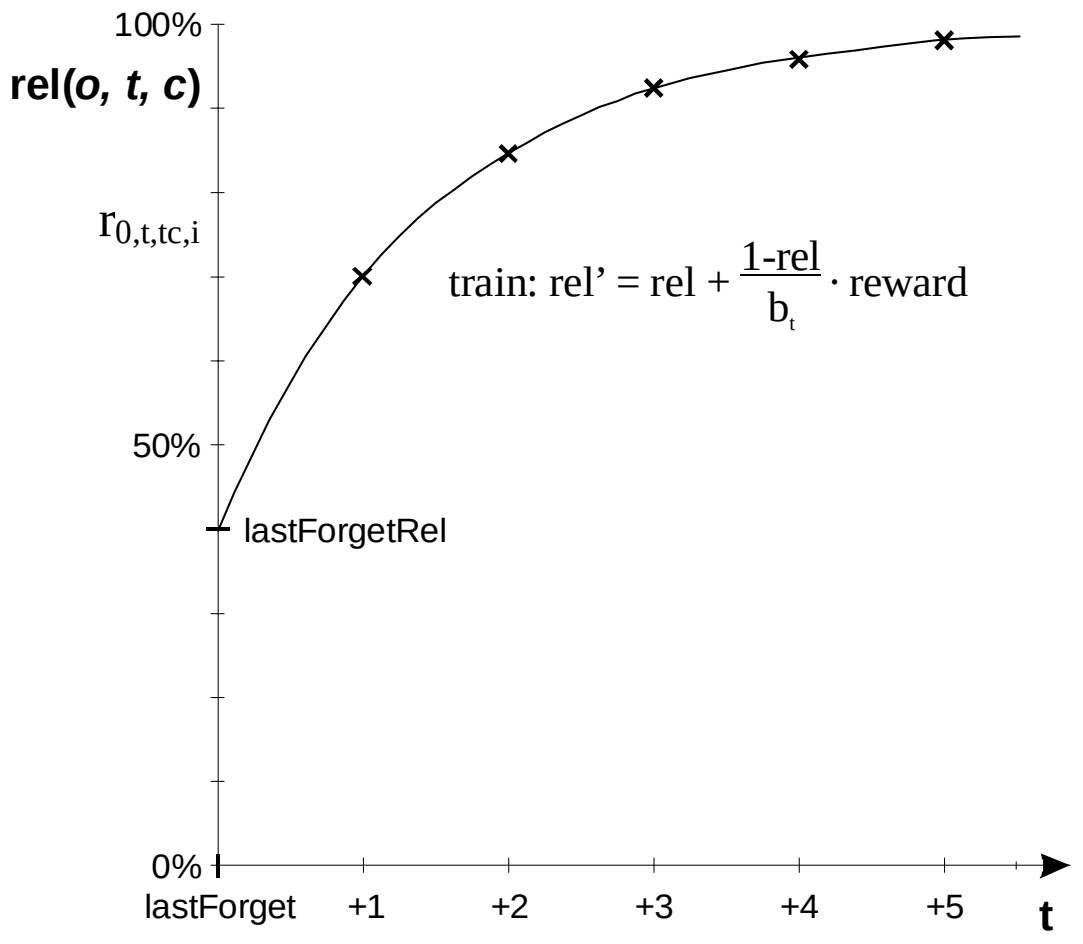

*Figure 4: Recursive train function with lastForgetRel(o, t, c)=0.4, reward=100% and $b_t$=2.0*

The following considerations show that this recursive function is equivalent to a power function: The rewards during a single training phase are constant for an object *o* in the context of a single task class *c* according to axiom 3 (section 2.3):

$\text{reward}(o, t, c) = r_{o\,c}$.

Additionally we use the following substitutions:

rel' := rel(*o*, *t*, *c*)  new relevance value

rel := rel(*o*, *t*-1, *c*)  last relevance value

This results in

$$\text{rel}' = \text{rel} + \frac{1 - \text{rel}}{b_t} \cdot r_{o,c} \text{ starting with lastForgetRel}(o, t, c) \quad [8]$$

$$\text{rel'} = \left(1 - \frac{r_{o,c}}{b_t}\right) \cdot \text{rel} + \frac{r_{o,c}}{b_t} \qquad [9]$$

with help of a z-transformation we get a non-recursive representation

$$r = \left(1-\frac{r_{o,c}}{b_t}\right)^t \cdot \text{lastForgetRel}(o, t, c) + 1 - \left(1-\frac{r_{o,c}}{b_t}\right)^t \qquad [10]$$

and finally

$$\text{train: rel}(o, t, c) = 1 + (\text{lastForgetRel}(o, t, c) - 1) \cdot \left(1-\frac{r_{o,c}}{b_t}\right)^t \qquad [11]$$

### 2.4.2 Function for Forgetting

The importance of *forgetting* for the performance of knowledge base search has been shown by several authors (e.g. [Markovitch et al. 1988], [Cox et al. 1992], [Oelesen et al. 1992], [Lin et al. 1994] ). Most of them use forgetting synonymously with the deletion of knowledge. Anderson has put forward that forgetting is a continuous process for humans (see section 1.3) and we are of the opinion that it should also be a continuous process in configuration domains since the deletion of knowledge is in this case mostly not adequate. For example the deletion of an old concept in a passenger car configuration domain would not only make the configuration of old timer cars impossible but also the reconfiguration of existing cars (which lies in our focus, see [Kreuz et al. 1999b]). The continuous process of ignoring old knowledge has also been used in aHUGIN [Olesen et al. 1992] and is called *fading* there: *"Variables in adaptation mode have an extra feature, fading, which makes them tend to ignore things they have learnt a long time ago, considering them as less relevant."* As a starting point for our forget-function we considered the power law of forgetting (see Section 1.3). The reason for this is that in technical domains similar characteristics are desirable for the development of relevance as in the human brain:

- new objects have an initial value of relevance which is much higher than old and unused objects. The information about the objects ages continuously
- initially objects lose their relevance faster. For example after 10 years, one week either way will no longer have a great effect on the relevance of objects in technical configuration domains
- the relevance approaches zero, whereas objects are never really irrelevant. The access just lasts longer

Real forgetting, i.e. the irreversible erasure of objects is not desirable at first. However in section 3 the issue is addressed as to how maintenance of a knowledge base can be supported with the help of relevance.

To make matters simpler the following assumptions were made compared with the function of the "power law of forgetting":

- reduction in the range of values to [0, 1]
- the relevance value during a forgetting phase should start at the last value of a preceding training phase lastUseRel($o, t, c$) $\in$ [0,1] and approach zero

The aforementioned "desired" characteristics remain. The first assumption is for the reason that we want to model relevance as percentage values. The second is because it is presumed that forgetting is a decay-process starting with the "*activation level*" after the last access.

If we try to determine the start relevance lastUseRel($o, t_0, c$) at the moment $t_0$ when an object $o$ is stored, we recognize that many parameters of the original function in the model of Cognitive Psychology are not available at first. For example in [Waltz 1999] it is described how the start relevance depends on the estimation of the importance of the information. In the best case the start relevance can be provided by the knowledge engineer who adds the object. If such an assessment is not available, we must assume that all objects that are added to a knowledge base are of the same importance because a computer cannot make "emotional" assessments. In this case we define a start relevance lastUseRel($o, t_0, c$) := 50% because with this relevance $o$ has the same chance to become relevant (to prove its worth) or to become irrelevant (to be forgotten).

***Definition 2:***

The start relevance for an object $o$ at the time $t_0$ when it is stored in context $c$ is set to 50%:

$$\text{start relevance: rel}(o, t_0, c) := 0.5 \qquad [12]$$

The basis $b_f$ of the power function is a parameter that is domain dependent, analog to the fact that some things are forgotten faster than others. The larger $b_f$ the faster is the decay process. The following definition describes the forget-function which we have applied:

***Definition 3:***

The process of gradually forgetting an object $o$ in technical domains can be described by the following function:

$$\text{forget: rel}(o, t, c) = \text{lastUseRel}(o, t, c) \cdot b_f^{-\text{age}} \qquad [13]$$

With

| | | |
|---|---|---|
| $\text{rel}(o, t, c) \in [0,1]$ | the relevance of an object *o* at time *t* during a forgetting phase in the context of task class *c* | [14] |
| age | $= t - \text{lastUse}(o, t, c) \in [0,\infty)$, time since the last use of *o* before *t* in the context of *c*, the time since the last training phase | [15] |
| $\text{lastUseRel}(o, t, c) \in [0, 1]$ | $= \text{rel}(o, \text{lastUse}(o, t, c), c)$, the relevance value after the last use of *o* | [16] |
| $\text{lastUse}(o, t, c)$ | the last time when *o* has been used before *t* in context of *c* | [17] |
| $b_f \in (1,\infty)$ | basis for the forget function determining the "speed" of forgetting. The larger $b_f$ the faster objects are forgotten | [18] |

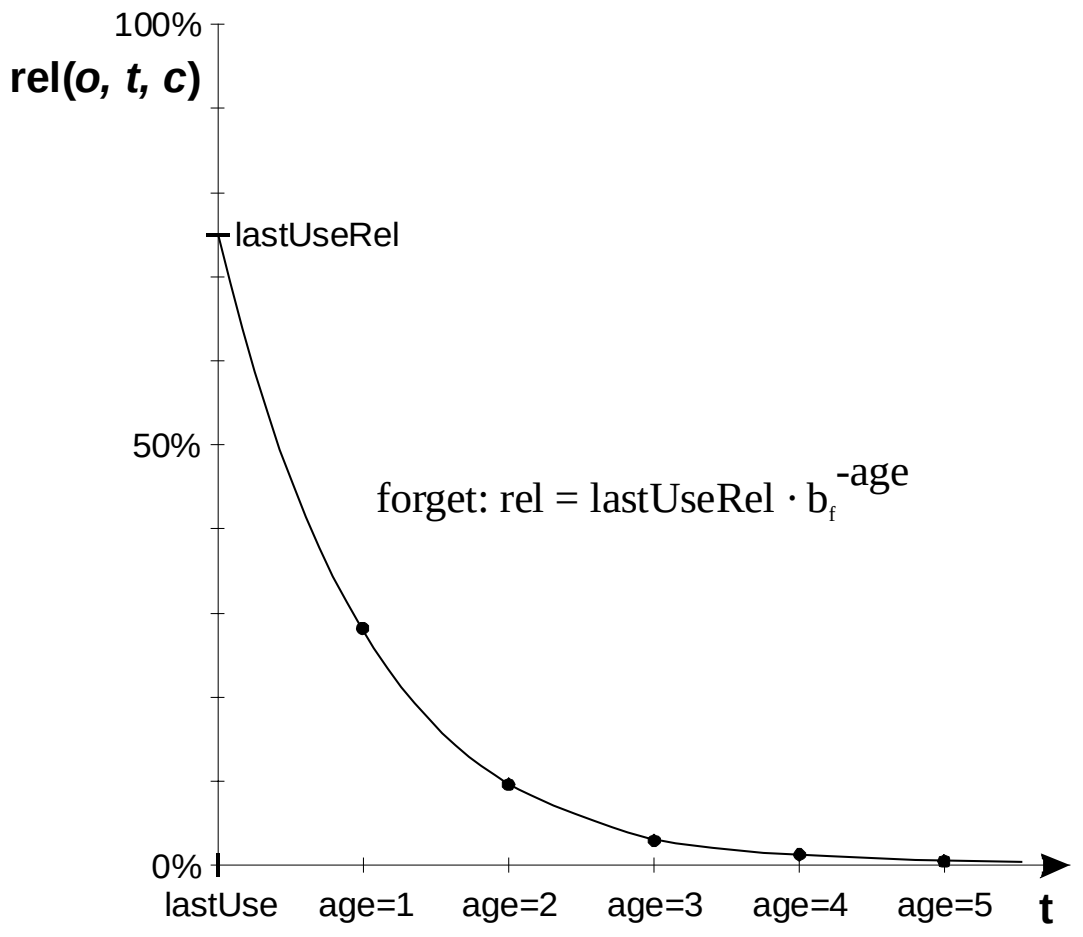

*Figure 5: The forget function for lastUseRel=75%, $b_f$ =* **e**

## 2.4.3 Relevance of Objects

In equation [2] the relevance function for an object *o* results from combining the forget and the train functions dependent on object *o* being part of the solution or not. With equations [3] and [13] the following definition results from [2]:

***Definition 4:***

The relevance of objects in a knowledge base is described by the following function:

$$\text{rel}(o, t, c) := \begin{cases} \text{rel}(o, t\text{-}1, c) + \dfrac{(1 - \text{rel}(o, t\text{-}1, c))}{b_t} \cdot \text{reward}(o, t, c) & \textbf{train}\text{, if } o \textbf{ is} \text{ part of the solution, start with lastForgetRel}(o, t, c) \\ \text{lastUseRel}(o, t, c) \cdot b_f^{-\text{age}(o, t, c)} & \textbf{forget}\text{, if } o \text{ is } \textbf{not} \text{ part of the solution} \end{cases} \qquad [19]$$

with

| | |
|---|---|
| rel($o$, $t$, $c$) $\in$ [0,1] | relevance for an object $o$ at time $t$ in the context of task class $c$ |
| reward($o$, $t$, $c$) $\in$ [0,1] | the user reward for object $o$ at time $t$ in context of $c$ (see [5]) |
| $b_t \in (1,\infty)$ | "basis" for the train function determining the "speed" of training. The larger $b_t$ the slower is training. Typical values lie between one and three (see [6]) |
| lastForgetRel($o$, $t$, $c$) $\in$ [0, 1] | last relevance value of $o$ before $t$ when the last forgetting phase ended in the context of the task class $c$ (see [7]) |
| age | = t – lastUse($o$, $t$, $c$) $\in [0,\infty)$ time since the last use of $o$ before $t$ in the context of $c$. This is the time since the last training phase (see [15]) |
| lastUseRel($o$, $t$, $c$) $\in$ [0, 1] | = rel($o$, lastUse($o$, $t$, $c$), $c$), the relevance value after the last use of $o$ (see[16]) |
| lastUseRel($o$, $t_0$, $c$) := 0.5 | the start relevance is set 50% when $o$ is stored (see [12]) |
| lastUse($o$, $t$, $c$) | the last time when $o$ has been used before $t$ in context of $c$ (see [17]) |
| $b_f \in (1,\infty)$ | basis for the forget function determining the "speed" of forgetting. The larger $b_f$ the faster objects are forgotten (see [18]) |

The following diagram shows an example of the relevance function from equation [19]:

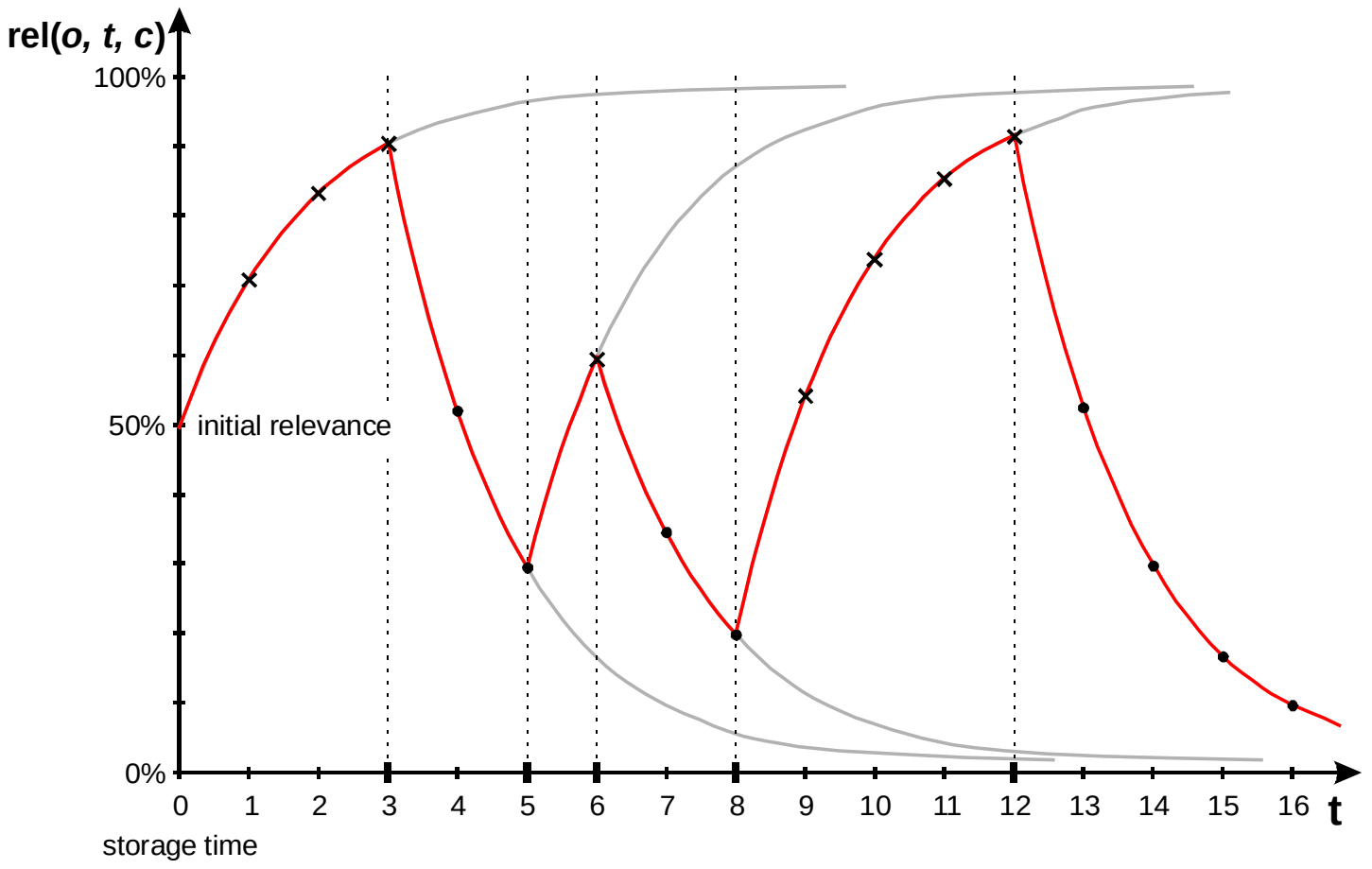

*Figure 6: Train and forget phases alternate in the relevance function dependent on object o being part of a solution or not*

In this example an object $o$ has been stored at $t$=0 and has been part of the solution at $t$={1, 2, 3, 6, 9, 10, 11, 12} with a constant user reward ("crosses" in Figure 6). According to that the intervals (0, 3], (5, 6] and (8, 12] are training phases. During the remaining intervals the object is continuously forgotten and from $t$=13 onwards it ages infinitely ("dots" in Figure 6).

# 3 Application of RKF in Configuration Domains

With equation [1] RKF will always help if one object has to be selected from a discrete and finite set of possible objects. During a configuration process in a domain that has been modeled with objects or frames, selections of that kind occur during the following configuration steps (with possible configuration steps we have oriented ourselves at PLAKON (see chapter 4.5 in [Cunis et al. 1991]):

**- specialization**

During a specialization step the "type" of a solution instance has to be specialized from an abstract concept to one of its children according to the taxonomical hierarchy. In this case the set of objects in question ***O*** is formed by the child concepts and *o* is one specialization to be selected.

**- decomposition**

A decomposition step determines the number of components of a concept. In many technical domains the set of possible numbers is discrete and each number can be understood as an object. Unfortunately in many examples no upper bound for the number of components can be given at first and consequently "infinite" (∞) is often used as an "upper bound" in the compositional hierarchies. Since RKF can only be applied to finite sets the determination of the number of components can not generally be supported by RKF. Nevertheless we were able to find an upper bound in all our domains by adding information from experts to the domain analogues to the following example: It's true that a PC can contain any number of hard disks in principle. But if we would ask an expert on PCs he or she would know that the maximum number of hard disks is probably given in a server PC, which "had never more than about 20 hard disks as far as he or she can remember". If we double this value we will probably not exclude any solution in that domain. The set of numbers [0..40] can be understood as a discrete and finite set of 41 objects and RKF can be used to select one number.

**- parameterization of instances**

With the use of RKF for parameters of instances it is very much the same as with the use for decompositions before: As long as the sets are discrete and finite RKF can be used. For example the set of colors of cars satisfies the presumption. But the value ranges of parameters like the size or position of an instance are not even discrete. So again we can say that RKF cannot help in principle to determine the parameters of instances. But nevertheless methods exist to reduce such sets to discrete finite sets, for example by dividing the value range into a discrete and finite set of intervals. RKF can help to select the right interval.

**- integration**

An integration step determines how many existing instances can be used to satisfy a relation between instances and how many concepts have to be instantiated. With the same limitations as with decomposition, these values can have a discrete and finite value range. How many concepts should be instantiated or integrated usually depends on the control knowledge and not on task classes. But RKF only uses a task class *c* as the context for learning and forgetting and will usually not help here.

**- agenda**

Configuration systems that use a structure based or object orientated representation typically use a so-called agenda to collect the configuration steps which have to be performed before the configuration is complete. Before each configuration step the configuration system has to determine which entry of the agenda should be processed next. The set of entries in the agenda is discrete and finite but no user assessments are available since the user is shielded against the control of a configurator.

**- heuristic selection**

The above list of possible applications shows that RKF, like almost all heuristics, can not be used in all situations optimally. In his doctoral thesis Günter [Günter 1991] has proposed the use of a flexible control that uses multiple heuristics dependent on so-called control knowledge. The set of heuristics is discrete and finite but again which heuristic to use usually is nothing that is typical for a task class and RKF will probably not help here. But if the heuristics to be used is typical for the task classes in a domain, RKF can be used as a meta heuristic that learns which heuristic helped most often to find a solution in the context of a task class. Anyway RKF is an additional heuristic that can be selected by the flexible control of a configuration system.

**- Maintenance of Knowledge Bases**

One important disadvantage of long existing databases and therefore also of knowledge bases is that objects which are no longer relevant cannot be automatically deleted. It is a conventional approach to delete *old* objects. With RKF this approach can be considerably improved by deleting not *old* but *no longer relevant* objects, thereby taking into consideration how often an object has been used “during the last time”. For this the relevance of all objects in the knowledge base could be determined and as soon as it falls short of a certain level of relevance the object can be deleted. For each object the highest relevance of all task classes is used, i.e. an object should only be deleted if it is no longer relevant in the context of all task classes.

# 4 Realization and Experiments

## 4.1 RKF-Configurator

RKF should be used together with other heuristics when used in a professional configuration system as already mentioned under “heuristic selection” in section 3. To be able to demonstrate the effects arising solely from RKF we have built up a configuration prototype (“RKF-Configurator”) in JAVA. We have already presented a first version of it on PUK2000 (see [Kreuz 2000]). It supports an object oriented representation of the configuration knowledge with concept hierarchies (taxonomical and compositional), n:m-relations and parameter relations (JAVA methods that return true or false) which we call “parameter constraints”. With the RKF-Configurator we have implemented a depth-first search with chronological backtracking. RKF is the only heuristic used and we have compared the results with “no heuristic used” which is equal to a random selection of objects during the depth-first search. RKF is used during all specialization and decomposition steps. The use of RKF for decomposition steps is always possible because we do not allow infinite (∞) as a number in the compositional hierarchies. The use of RKF for the parameterization of instances has been planned but we do not expect effects different to that arising from the use in selecting numbers during decomposition steps. Consequently we have not implemented a complete support for parameterization and parameters are only collected along the taxonomical hierarchy instead. The configuration is built up using specialization, decomposition and “parameterization”. Subsequently the n:m-relations and parameter constraints are tested at the end of such configuration steps when the configuration is possibly complete and might cause backtracking (build and test). In our simple implementation we assume that *all* n:m-relations can be realized by integrating the existent instances. So we do not need to support integration steps by RKF. Since we use depth-first search and chronological backtracking as the only strategies no selection from the agenda can be supported by RKF and because RKF is the only heuristic it is also not used as a meta-heuristic.

To start a configuration a user can select a concept from the compositional hierarchy as the root of the system to be configured and the task class as the context for the actual configuration task. RKF needs no further user input during the configuration process and next the configurator will present a solution if possible. In this case the user is prompted to give an assessment for every single component of the solution. In the case that the configuration found is still not good enough relating to the above assessment the configuration can be restarted. Due to the probabilistic selection method in RKF the same task specification can lead to different (but similar) solutions. In our project at this point we export the solution as a XML-file, the so-called *Exact Configuration Onboard* which is a documentation of a vehicle that is stored onboard this vehicle (see [Kreuz et al. 1999a], [Kreuz et al. 1999c], [Kreuz et al. 2000] ).

## 4.2 Implementation Considerations

The most important extension to a configuration system that should use RKF is to store the relevance information for each object that should be selectable with the heuristic RKF. Relevance information consists of two elements:

1. lastUse(*o*, *t*, *c*)  the point in time when *o* has been used lastly

2. lastUseRel(*o*, *t*, *c*) the relevance value which *o* had at that point in time

With that information the actual relevance of an object can always be calculated during a configuration process by using the “forget” portion of the relevance formula [19]: If *o* was **not** part of the solution in the configuration run before, this is obviously the correct part to be used. If *o* **was** part of the solution age(*o, t, c*) will return 1 which results in the proper value in that case, because one time unit lies between the actual time and the last use.

As soon as a configuration run has finished, only the relevance information of those objects has to be updated that were part of the solution (which are typically few relative to the total number of objects stored in a knowledge base). The relevance information of all other objects does not need to be updated: The “forget”-part of the relevance formula will provide the actual relevance value *if required* as just described.

To update the relevance information of objects that were part of the solution the actual relevance value is calculated and used as rel(*o, t*-1, *c*) in the “train” part of [19]. The result of this calculation is stored as lastUseRel(*o, t, c*) and *t* as lastUse(*o, t, c*). Not until now, when all necessary relevance information has been updated, the time *t* of the actual task class is updated (increased).

With this method only two additional elements are needed per object of the knowledge base for every task class. If $n=|\boldsymbol{O}|$ is the number of objects stored in a knowledge base and $k=|\boldsymbol{C}|$ is the number of task classes of a domain then a linear amount of $2k \cdot n$ of additional memory is needed.

## 4.3 Experiments

### 4.3.1 Example Domains and Conditions for the Experiments

We have successfully tested RKF in five very different technical domains. Although all of them do not consist of more than 127 concepts the number of possible combinations of these concepts (search space) is already enormous (see “possible combinations” in table 1). Without the help of heuristics the whole search space had to be scanned in the worst case scenario to find any or a even a “good” solution. Dependent on the nesting of a domain our implementation is able to compute (“build and test”) about 3,000 combinations per second on an 850 MHz PC. As an example about $323 \cdot 10^{12}$ years ($10^{12}$ = trillion = million million) would be necessary to scan the complete PC Domain (2.) with 125 concepts. To come to the point RKF has helped in all domains to find a good solution quickly by learning which objects can be used (only these get rewards as a part of the solution) and what are “good” solutions in respect to the user rewards.

| domain | number of concepts | possible combinations | remarks |
|---|---|---|---|
| 1. Simple ABC Domain | 26 | s1: 1,872<br>s2: 18,162,144<br>s3: 9,702<br>s4: 91,327,950 | we have used this very simple domain with 4 possible system roots to study the complexity of configuration problems |
| 2. PC Domain | 123<br>125 | $4.65 \cdot 10^{24}$<br>$21.22 \cdot 10^{24}$ | a typical domain for RKF use with short innovation cycles |
| 3. Simple PC Domain | 20<br>22 | 192,024<br>801,864 | we have used this simplified version of the PC Domain for extended experiments since all combinations can completely be calculated in 80 seconds even with wrong parameters for the RKF function |
| 4. Backrest Drive | 46<br>48 | 36,400,428<br>44,489,412 | prototypical example domain for an actual project at the University of Stuttgart |
| 5. COSIMA | 127 | $2.95 \cdot 10^{298}$ | prototypical example domain for an actual project at DaimlerChrysler R&T |

*table 1: Complexity of the domains in which we have applied RKF*

Most of the domains have been investigated with a basic number of concepts and with "two additional" concepts. The change of a domain should cause RKF to "retrain".

We will now introduce some of our experiments which we have performed with the "Simple PC Domain". The goals of the experiments were to proof that it is possible to *learn* with RKF which concepts should be used to get high rewards and that RKF can react to changes in a domain ("retrain"). We have also investigated the effects arising from different values for the parameters of RKF ($b_t$, $b_f$ and $v$).

We will first introduce the domain briefly:

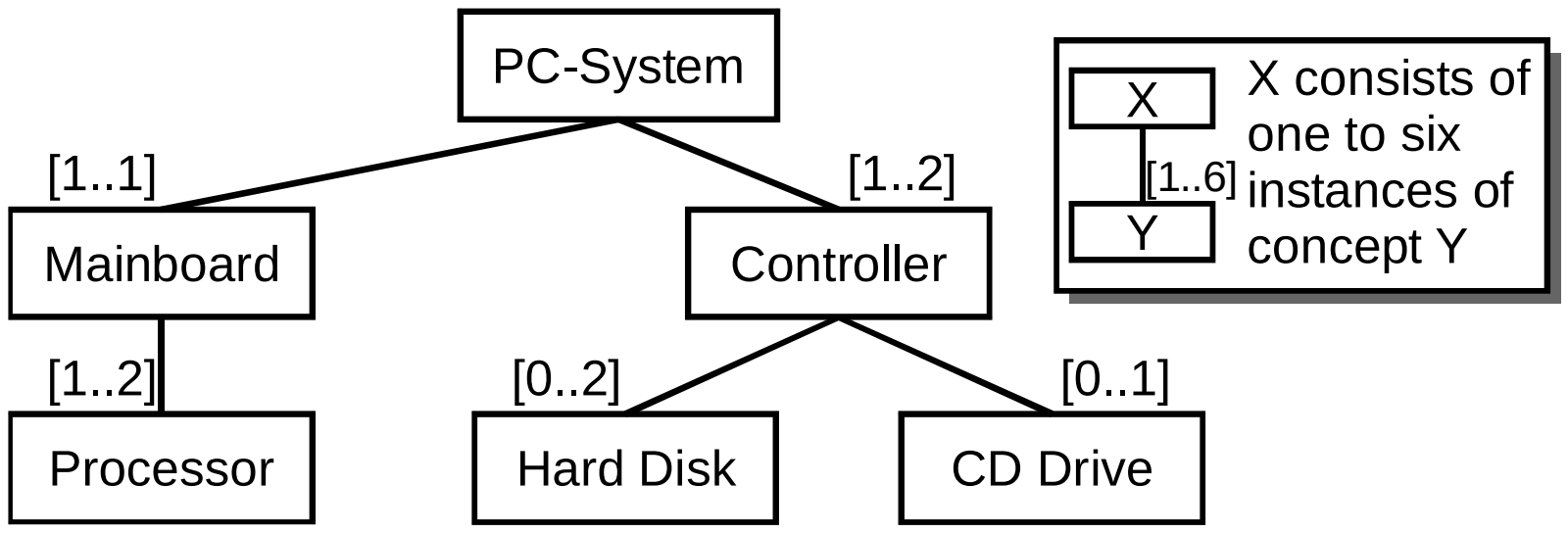

*Figure 7: Compositions of the "Simple PC Domain" example*

In this simplified version of the PC-Domain a PC-System consists of exactly one mainboard, which again consists of one or two processors, and one or two controllers, which again consist of up to two hard disks and one CD-Drive optionally. The following specialisations can be used for the concepts:

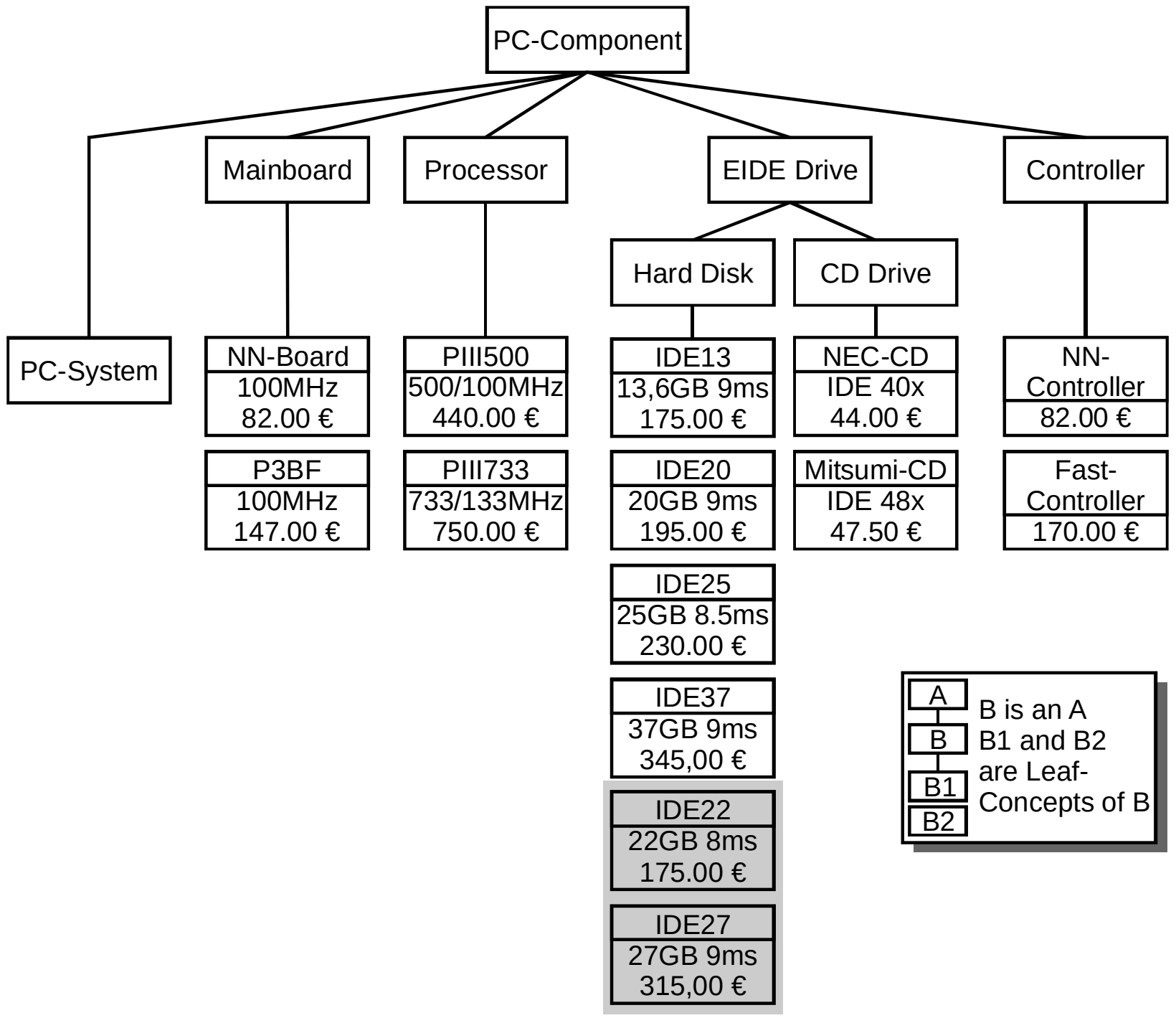

*Figure 8: Taxonomy "Simple PC Domain"*

The concepts "IDE22" and "IDE27" are used to change the domain and they are added during the experiments. Two n:m relations exist in the domain (see Figure 9) which cause that a cheap NN-Board must only be used together with cheap NN-Controllers and the more expensive P3BF mainboard must only be used together with the faster processor.

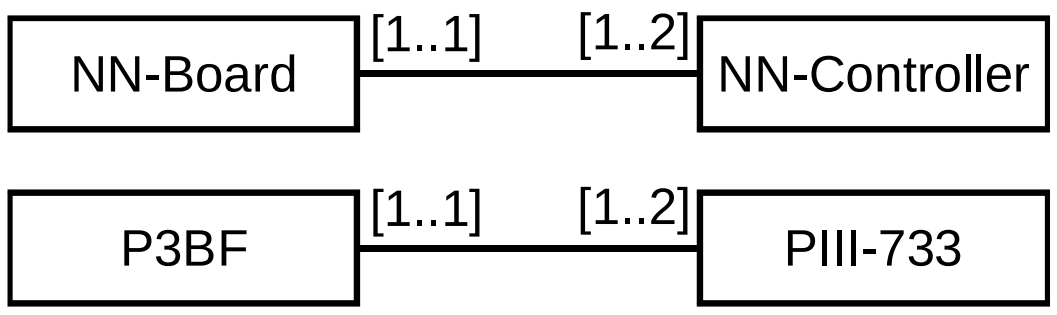

*Figure 9: Two n:m relations exist in "Simple PC Domain"*

Our configurator tries to fulfill the relations by integration only ("test") after a combination has been built on the basis of the composition and taxonomy ("build"). Consequently 40,068 of the 192,024 possible combinations (four hard disks) and 167,184 of the 801,864 possible combinations (six hard disks) become invalid and will cause backtracking. RKF should learn how to choose concepts to avoid backtracking.

All experiments were performed in the context of the same task class "Home-PC". To make possible an unattended batch operation, the rewards are given automatically according to table 2 and table 3.

| concept | reward | | implicit specification to be learnt in context of class “Home-PC” |
|---|---|---|---|
| | $t < 100$ | $t \geq 100$ | |
| NN-Board<br>P3BF | 100%<br>30% | | cheapest mainboard |
| PIII-500<br>PIII-733 | 100%<br>30% | | cheapest processor |
| IDE13<br>IDE20<br>IDE25<br>IDE37<br>IDE22<br>IDE27 | 100%<br>10%<br>10%<br>50% | 1%<br>10%<br>10%<br>10%<br>100%<br>5% | cheapest or maybe largest hard disk. After addition of the two new hard disks, IDE22 should dissolve the so long most relevant |
| NEC-CD<br>Mitsumi-CD | 20%<br>100% | | cheapest CD drive |
| NN-Controller<br>Fast-Controller | 100%<br>20% | | cheapest controller |

*table 2: Rewards for concepts in context of the task class “Home PC”*

| content relation | reward | implicit specification to be learnt |
|---|---|---|
| PC-System - - - [1..1] Mainboard | 1: 100% | |
| PC-System - - - [1..2] Controller | 1: 100%<br>2: 100% | what ever is allowed is ok |
| Mainboard - - - [1..2] Processor | 1: 100%<br>2: 10% | a Home PC consists of one processor |
| Controller - - - [0..2] Harddisk | 0: 0%<br>1: 100%<br>2: 10% | normally one hard disk, two are not really wanted and a diskless PC is absolutely not preferred |
| Controller - - - [0..1] CD-Drive | 0: 10%<br>1: 100% | one CD-ROM is recommended for Home PCs |

*table 3: Rewards for the number of contents in context of “Home PC”*

All experiments consisted of 200 configuration runs. Halfway through each experiment ($t$=100) the two hard disks IDE22 and IDE27 have been added and the table of rewards has been altered. Each experiment with 200 steps has been repeated for at least 10 times to minimize the possibility of “by chance” results.

### 4.3.2 Results

First we have investigated the effect of the constants $b_t$, $b_f$ and $v$, i.e. the effect from faster or slower forgetting and training as well as the control of conservatism in selections. The curve of the relevance values of the six hard disks were of special interest to us because here we applied the change to the domain and we can observe how training and retraining works.

In our first experiment we have used the basis $b_f = b_t = 2.0$ like we have used them in section 2.4 deriving the relevance formula. With $v = 1.0$ we have selected a linear interpretation of relevance during selections. Figure 10 shows a typical result for these values. The diagram shows the probability with which a hard disk concept is chosen according to its relevance (see equation [1]).

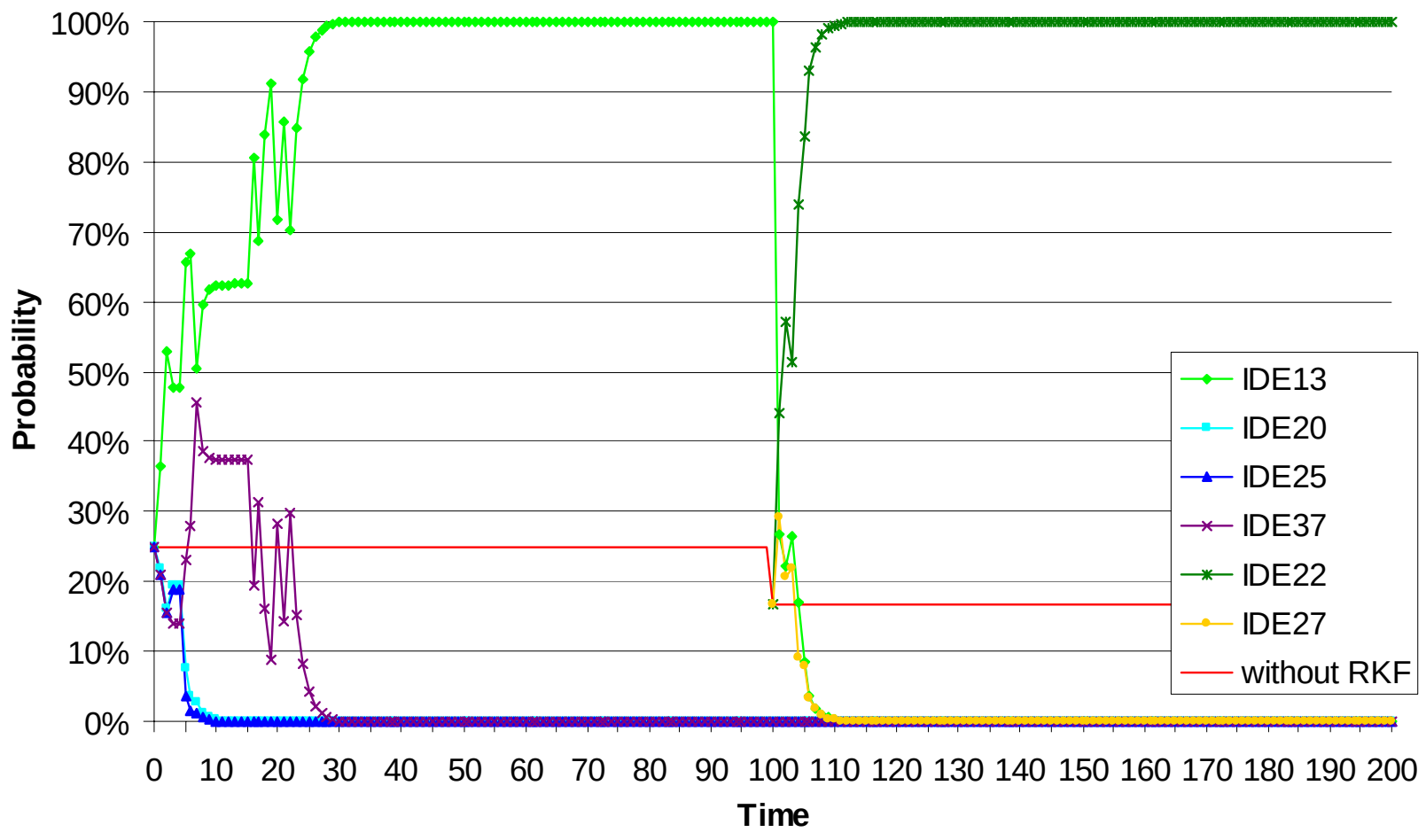

*Figure 10: IDE13 has been learnt as the concept with highest rewards correctly ($b_f$ = $b_t$ = 2.0 and v = 1.0)*

At first the probability of choosing one of the four hard-disk-concepts is 25%, because initially their relevance is equally the start relevance of 50%. After a short transient oscillation IDE13 “wins” because of a permanent high rating of 100%. The values of $b_f$ and $b_t$ make learning and forgetting so quick that other concepts are no longer chosen after 30 configuration runs. As soon as the two new hard disks are added (*t* = 100) the relevance (and with it the probability) of IDE13 decreases as a consequence of fast forgetting and slower learning due to low rewards. The new concept IDE22 can take over from IDE13 after less than 10 runs. With their start relevance of 0.5 the new hard disks are the only concepts that can compete with IDE13 and although IDE13 is selected once more (t=103), it is out done by IDE22 quickly because of its low reward and the high reward of IDE22. IDE27 had only be chosen twice during the experiment (*t*=101 and t=103) because of bad rewards following these choices.

These results appear to show the exact behavior for which we had desired: The concept with the best rewards is chosen with highest probability after a short learning phase and “forgotten” as soon as a better concept exists. Unfortunately not all of the experiments we have made with these constants show the same results. Since we use a probabilistic method it is possible that IDE37 is chosen at the beginning when all concepts have the same relevance and since it gets a fairly good reward it becomes quite a bit more relevant and therefore more probable for the next run. All the others, including IDE13, are becoming less relevant because they were not part of the solution. If training and forgetting is too fast RKF will trust such a “early starter”: In five of ten experiments with these parameters IDE37 has become most relevant instead of IDE13, one of them is shown in Figure 11.

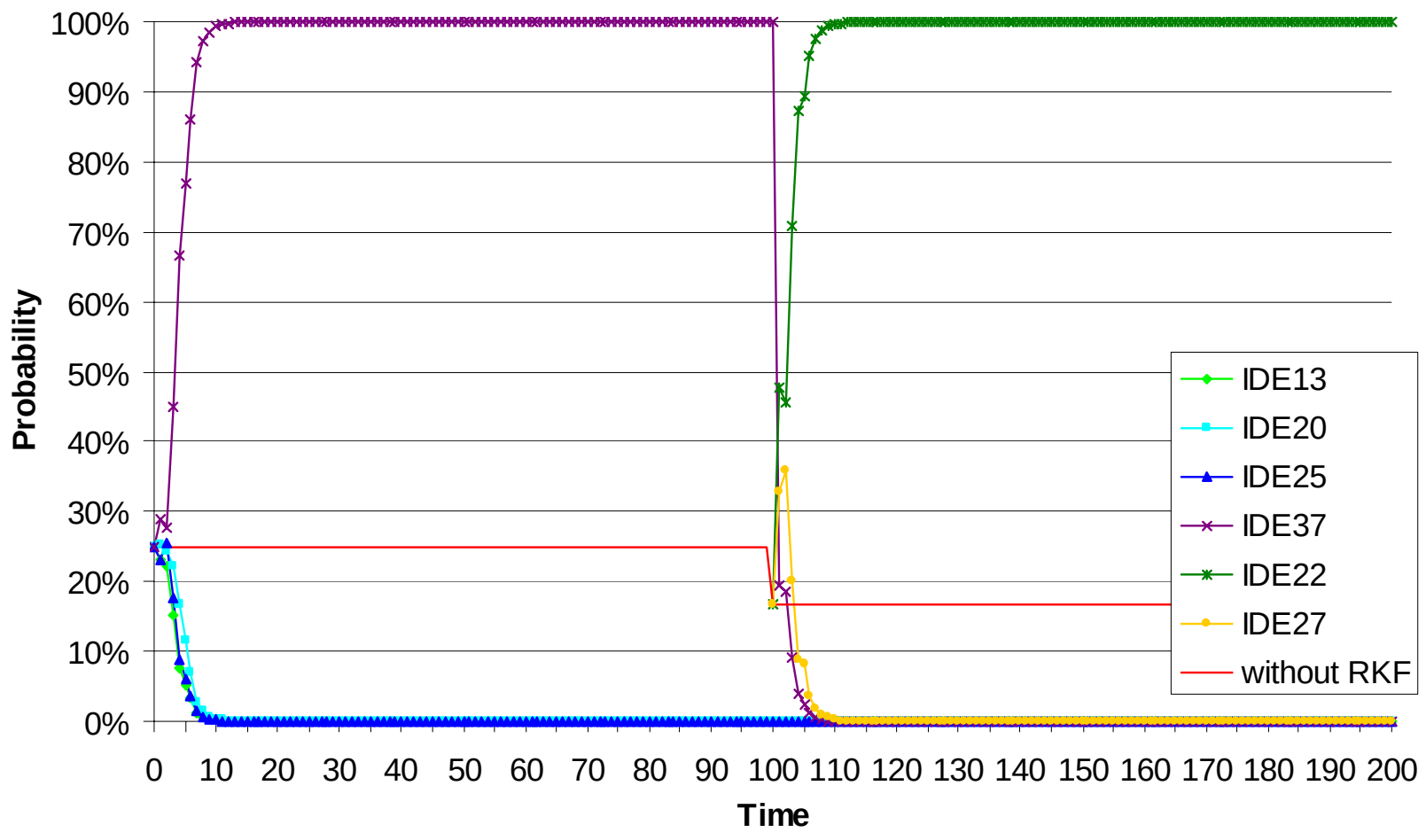

*Figure 11: Sometimes IDE37 "wins" errorneously using $b_f = b_t = 2.0$ and $v = 1.0$*

During our experiments with different parameters $b_f$, $b_t$ and $v$ we could validate the effects which we have listed in table 4, table 5 and table 6.

| | |
|---|---|
| *t1* | the speed of training determines how fast the relevance of an object approaches **100%** that is part of the solution repeatedly |
| *t2* | larger $b_t$ → slower training<br>smaller $b_t$ → faster training |
| *t3* | since the user rewards are multiplied with $1/b_t$ a large value has also the effect that the rewards are not considered so strongly additionally to slower training |
| *t4* | the **smallest** possible value of 1.0 together with a user reward of 100% would cause the relevance of an object to become 100% in just one step! |
| *t5* | the **largest** value we have used was 10.0 which made the influence of the rewards already so little, that we could not observe any learning that would reflect the rewards given (see equation [3]) |

*table 4: Statements about the effects arising from different values for* $\mathbf{b_t}$

| | |
|---|---|
| *f1* | the speed of forgetting determines how fast the relevance of the objects approach **0%** that were not part of the solution |
| *f2* | larger $b_f$ → faster forgetting<br>smaller $b_f$ → slower forgetting |
| *f3* | the **smallest** possible value of 1.0 makes forgetting infinitely slow and switches off forgetting! |
| *f4* | the **largest** value we have used was 10.0 which made forgetting so fast, that all objects that were not chosen at the first experiment had a relevance of near zero immediately and have not been chosen anymore. Since all objects in the first experiment have their start relevance they are chosen by random and consequently it is also random which object wins |

*table 5: Statements about the effects arising from different values for* $\mathbf{b_f}$

| | |
|---|---|
| *v1* | being "more conservative" during a selection means that an object that has been selected before is selected again more probably, because of having a slightly higher relevance value |
| *v2* | If an object is selected repeatedly it gets rewards more often and consequently its relevance reaches 100% quicker |
| *v3* | larger → more conservative → faster training<br>smaller → less conservative → slower training |
| *v4* | the **smallest** possible value of 1.0 causes probability that is linear to the relative relevance of an object |
| *v5* | values **larger** than 2.5 makes the selection so conservative that it becomes impossible to dissolve an object with a new one |

*table 6: Statements about the effects arising from different values for **v***

We can solve the above problem with "wrong" objects becoming the most relevant by slowing down the speed of training and forgetting by adjusting the three parameters (large $b_t$, small $b_f$ and small $v$). By doing this the transient oscillation lasts longer and it becomes more probable that the concept that really gets the best rewards is chosen during that time and consequently will win. The price we pay is a longer period with possibly suboptimal results.

With this in mind we have performed a series of experiments with different parameter sets. We wanted to find a combination that allows the configurator to learn which objects do not get good rewards (≤ 10%) within the first few steps (<10). The probability of IDE37 with a reward of 50% and IDE13 with the highest reward of 100% should oscillate but IDE13 finally should "win" (have a higher probability than 70%) before t=100.

To take the rewards strongly into consideration we have chosen a small value of 1.4 for $b_t$ according to statement *t3* (see table 4). We have chosen a small value for $b_f$ which causes slow forgetting to get a longer oscillation phase. We started with $b_f$=2.0, gradually decreased the value and finally used $b_f$=1.1. This value causes the oscillation phase to be noticeable longer than 100 steps. Finally we have increased $v$ gradually, starting from v=1.0 to make the selection slightly more conservative and to shorten the oscillation phase again. A value of $v$=1.9 was the largest which still allowed IDE22 to dissolve IDE13 in all experiments reducing the oscillation phase to 70 steps in the worst case scenario.

Figure 13 and Figure 12 show the results of the experiments with these parameters. The curves show the average values of 10 subsequent experiments; but also all the single experiments showed the desired characteristics.

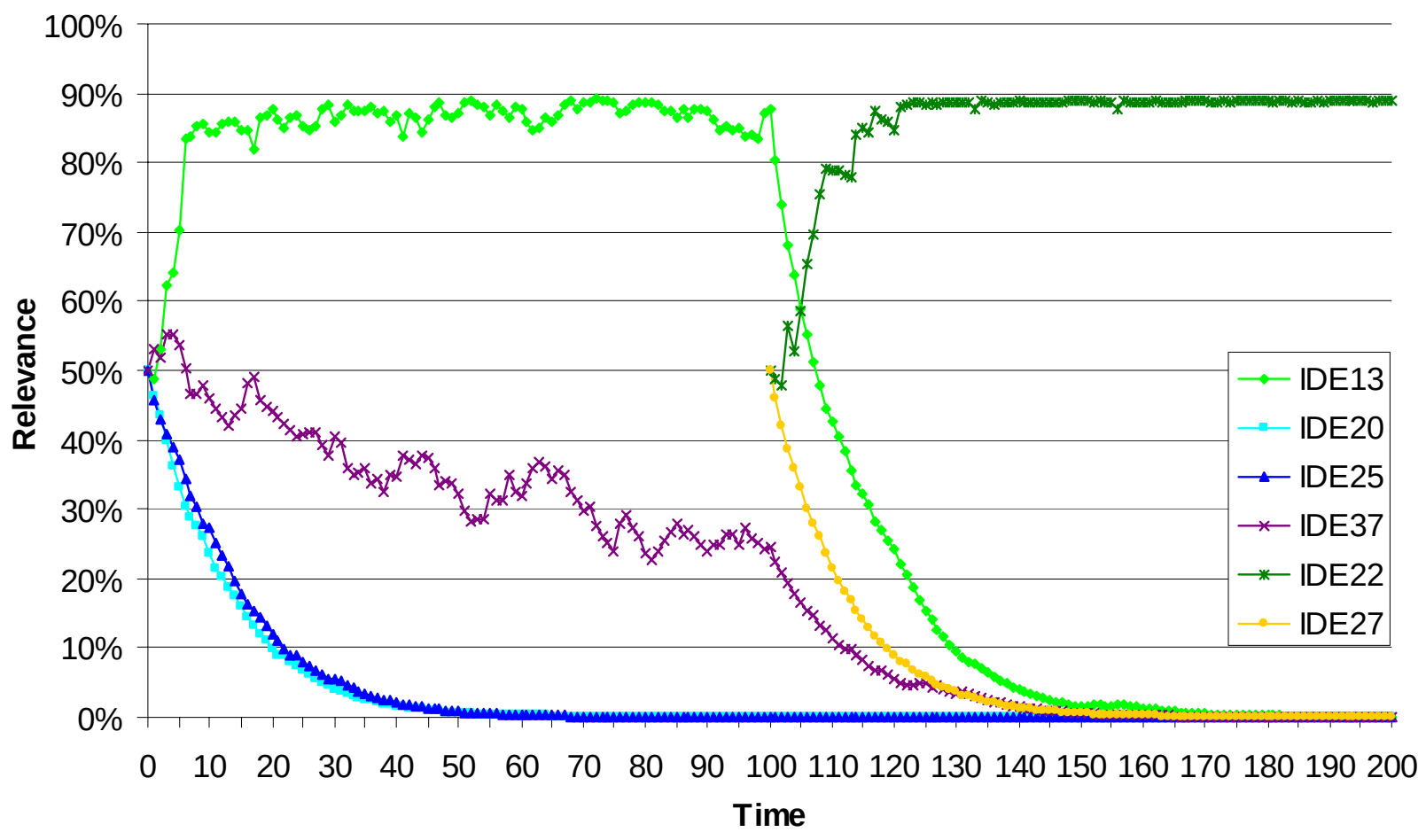

*Figure 12: Relevance values (average of 10 subsequent experiments) with parameters $b_t = 1.4$, $b_f = 1.1$, $v = 1.9$*

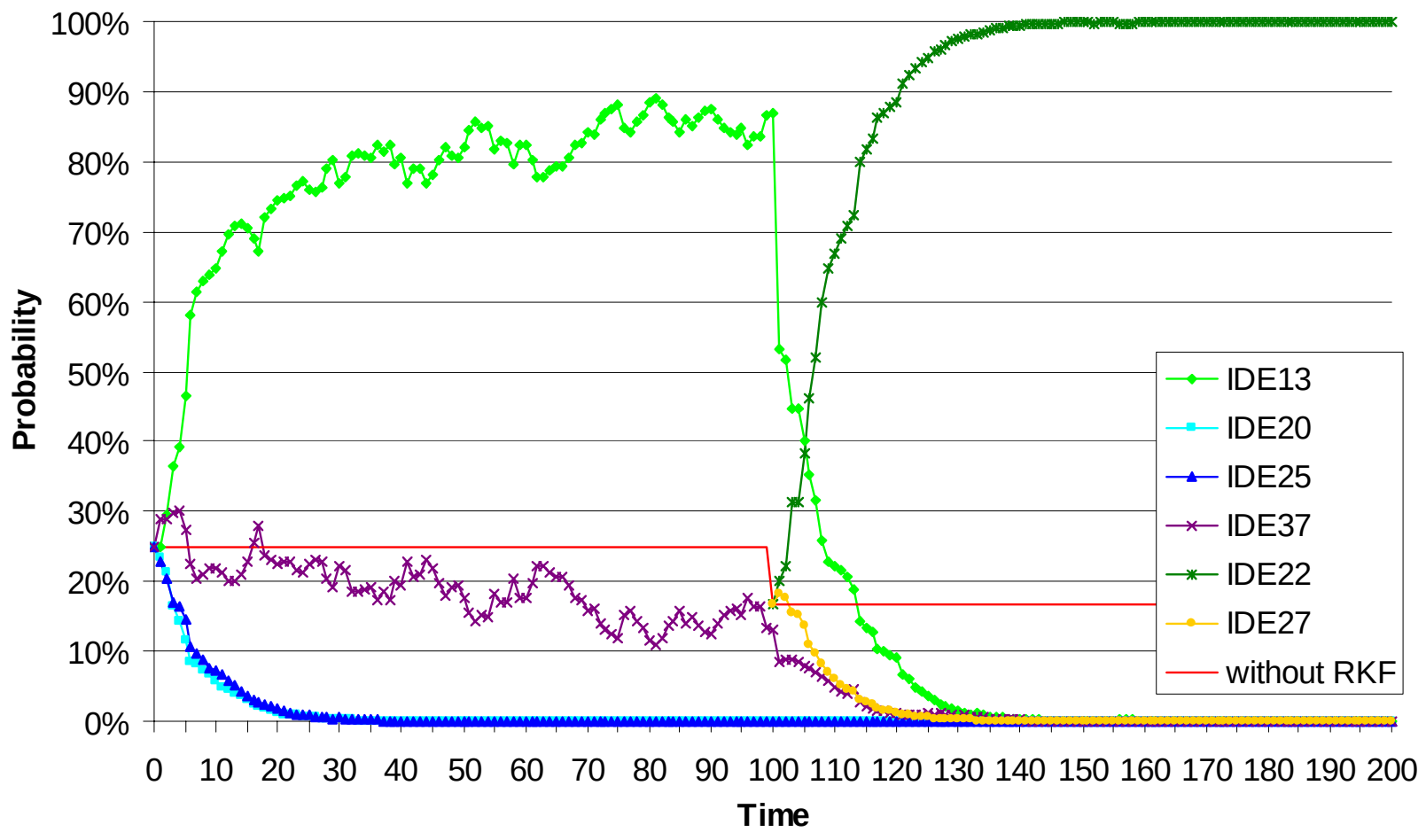

*Figure 13: Average probability values from 10 subsequent experiments with parameters $b_t = 1.4$, $b_f = 1.1$, $v = 1.9$*

The red line in Figure 13 shows the probability for the drives if they were chosen by random, i.e. without RKF. RKF should not only learn which concepts receive high user reward but also which combinations are allowed (see relations shown in Figure 9). Only the concepts will get rewards that are part of an (allowed) solution and finally only these concepts will be chosen by RKF. Figure 14 shows how many backtracking steps have been needed. The long oscillation phase is also reflected here and after about 65 runs RKF only selects concepts of an allowed combination: In the following configuration runs backtracking is needed anymore which also means that in a well trained system a solution can be found in a fraction of a second.

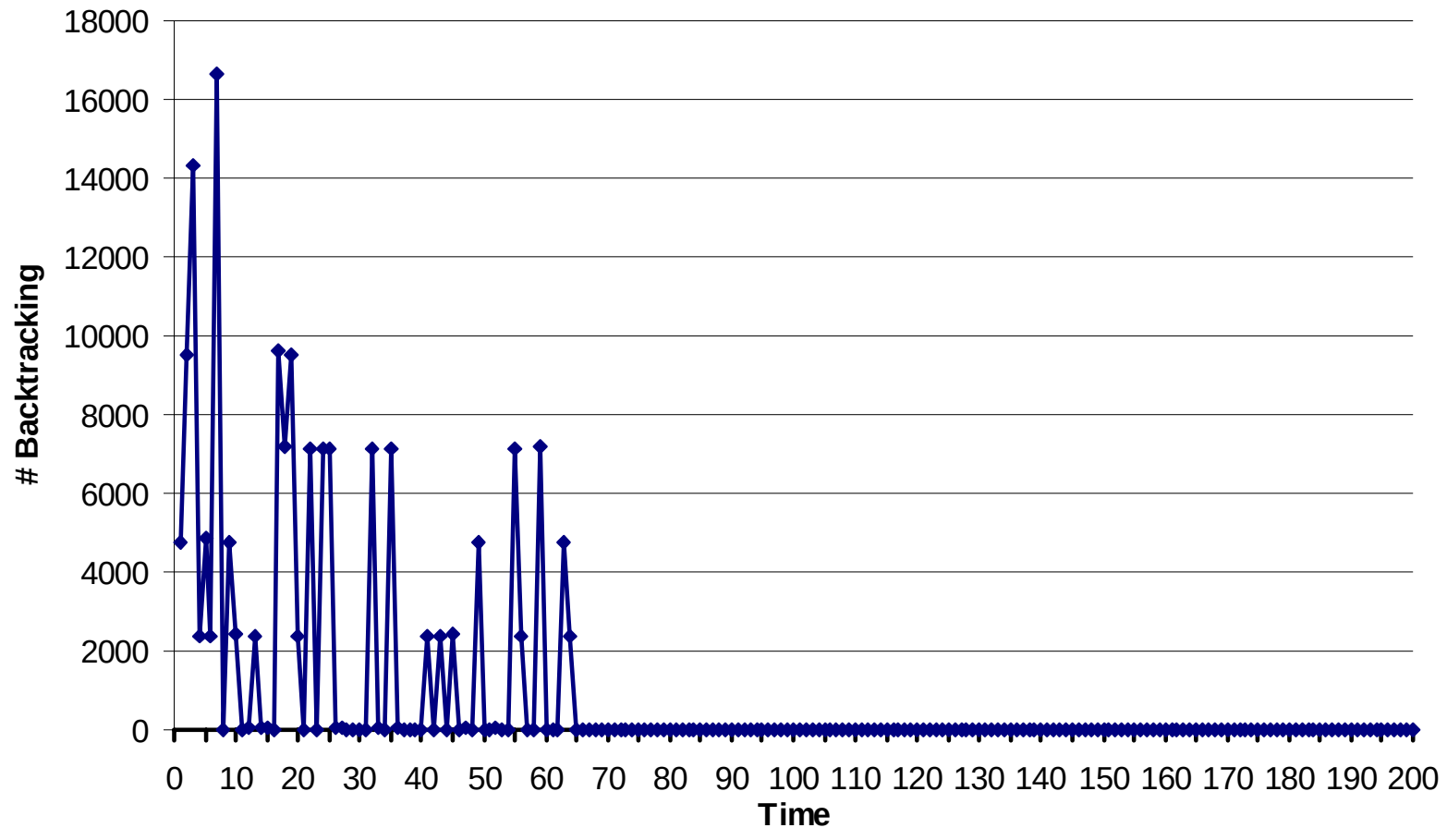

*Figure 14: Number of backtracking steps needed in "Simple PC Domain" (average of 10 subsequent experiments) with $b_t$:=1,4; $b_f$:=1,1; v:=1,9*

One question we had was "what happens without forgetting". To switch off forgetting we can simply set $b_f$ := 1.0. The results are shown in Figure 15.

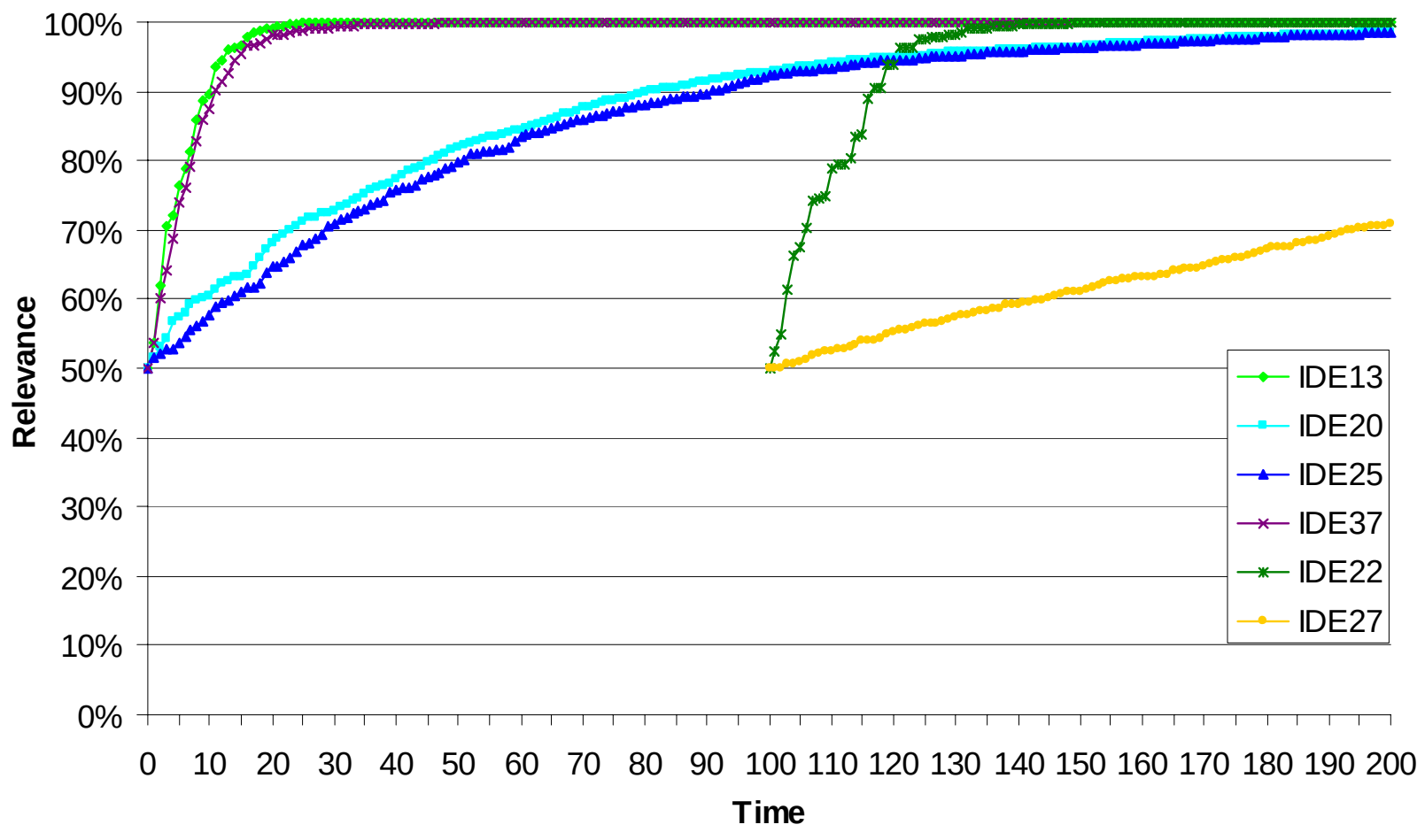

*Figure 15: Average relevance values without forgetting ($b_t$:=2,0; **$b_f$:=1,0**; v:=1,0)*

The relevance of *all* hard disks approach 100% because without forgetting the relevance function is monotonously growing. As a consequence the probability of a concept to be chosen differs only at the beginning, when high rewards made the "good" hard disks relevant more quickly.

After a short time the probability of all concepts approaches the value "without RKF" (see Figure 16). Other methods that use reinforcement learning without forgetting can do so because they do not use an exponential function that approaches any value. They learn infinitely instead (for example see [Kaelbling et al. 1996]) mostly using a linear learning function. With this they will only have the problem that they cannot react to changes in the knowledge base.

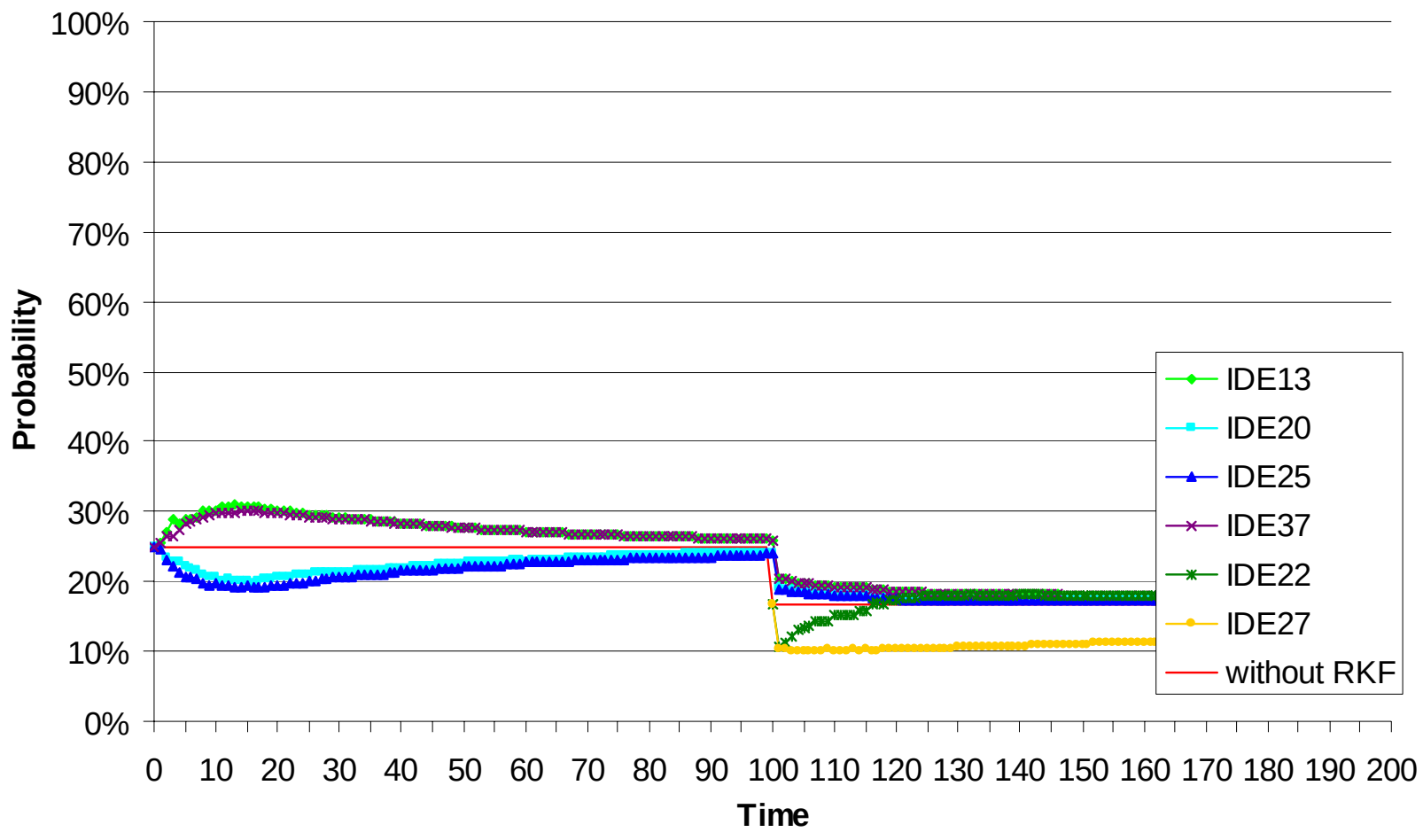

*Figure 16: Average probability values without forgetting*
*($b_t$:=2,0; **$b_f$:=1,0**; v:=1,0)*

## 4.4 Effects arising from RKF

With our experiments we could confirm the following effects:

- *Increase in Efficiency:*

RKF supports the search for solutions by helping to make correct decisions and thereby avoiding backtracking.

Through the application of RKF a knowledge-based system "learns" from previous successes.

The search space is not really restricted with RKF, however the paths in the search tree are favored which have often led to a solution. Figure 17 shows part of the taxonomy of the PC Domain (example 2 in table 1).

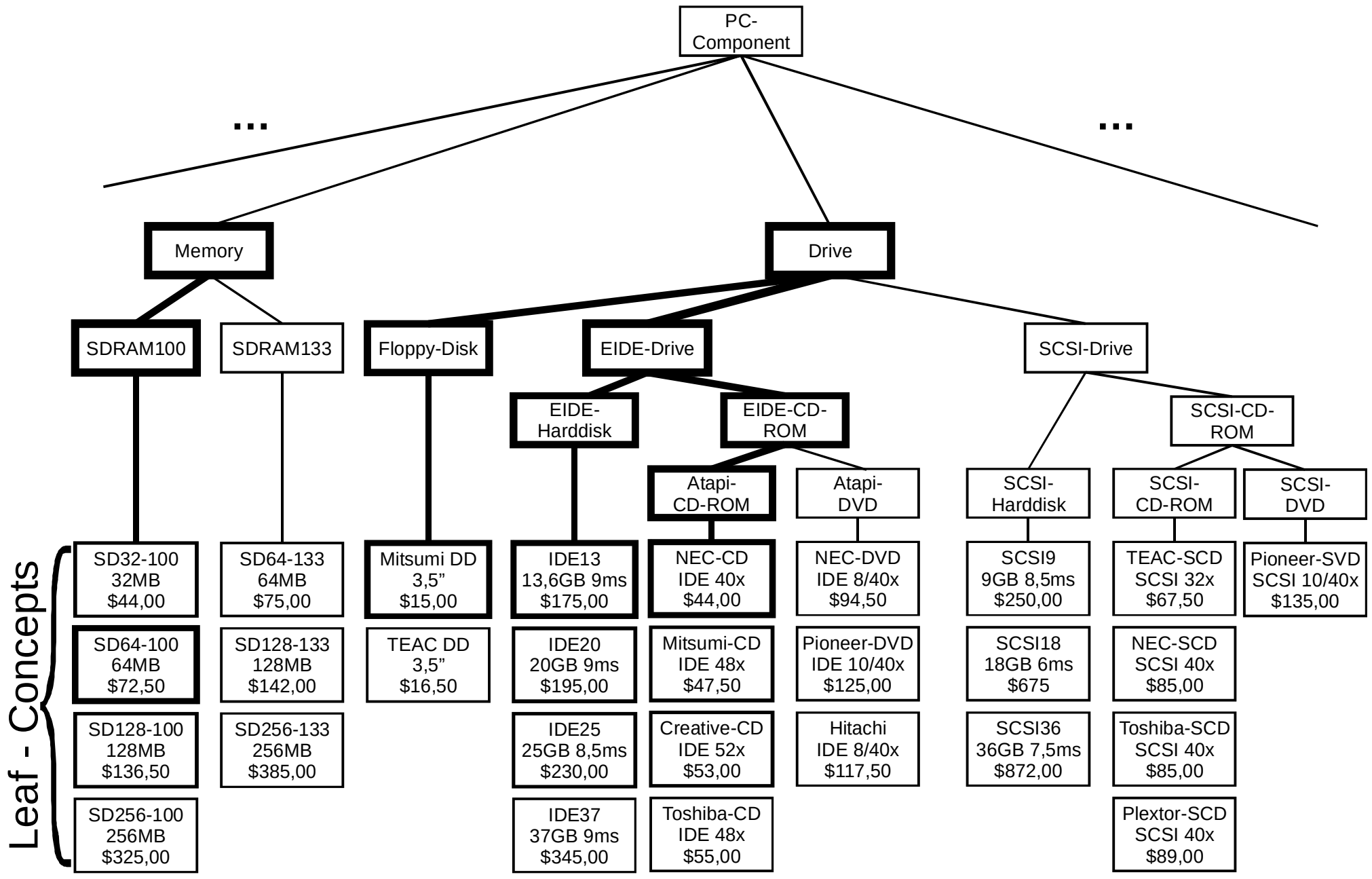

*Figure 17. Part of the taxonomy of the PC Domain. The thickness of the lines indicate the relevance for the task class "Home-PC"*

The thickness of the lines indicates the relevance learnt for the task class “Home PC” i.e. this form of PC most often contains a SD64-100 memory module, a Mitsumi DD disc drive, an IDE13- hard drive and a NEC-CD drive. The figure also shows how RKF supports the depth first search in this hierarchy: Paths of high relevance develop because whenever a daughter concept was useful, the respective father concept was also useful. This is the main reason why configuration with RKF is so fast: If a specialization of a concept has to be chosen during the configuration process the RKF heuristic will first search along the relevance paths with high probability.

- *“Soft” versioning management*

Newer objects and therefore also objects of newer versions are preferred by RKF and so these are mainly used. The old versions can remain in the knowledge base. What is particularly pleasing about this procedure is that newer versions are automatically no longer used if they do not prove to be good and older versions are simply fallen back on. On the other hand, if a new version works well, an old version is used less. The old version however still remains available for “exotic” cases.

- *Task Classes do not age*

RKF has some similarities with case-based methods (CBR) with regard to the influences of previous solutions. An important advantage of task classes over cases is that they do not age. For example a “Home-PC” was completely different 10 years ago to today. A “representative” case from that time is completely useless today. The task class “Home-PC Class” however is unchanged because the problem, optimization objectives, the expected cost etc. stay the same.

- *Not Conservative*

Through preference to *new* objects, RKF is a heuristic which is not conservative despite learning from preceding solutions: New objects have the opportunity to “prove their worth” and they become irrelevant if they could not receive good rewards. This characteristic is certainly one of the most important advantages of RKF over methods which likewise include preceding solutions, e.g. pure reinforcement learning methods, as these are conservative as a rule.

## 4.5 Limitations

As well as the positive results already described before, our experiments also confirmed the weak points we had expected with the use of RKF.

- *at least n non-optimal runs*

RKF shares this problem with all reinforcement learning methods: If an object has to be chosen from a set of $n$ objects at least $n$ runs are necessary to get rewards for all of the objects. In RKF the parameters $b_t$, $b_f$ and $v$ have to be adjusted in a way that the oscillation phase lasts at least $n$ steps. During these steps non optimal results are gained. Probably domains exist where the largest *n* cannot be estimated in advance. Another possibility is not to begin with a start relevance of 0.5 but with a relevance equal to a first representative user reward for each object (if such values are available). In a trained system the value of *n* is not so important: Only one object per selection set can stay relevant ("will win") finally. The oscillation phase then needs only to be long enough to get rewards for *m* new objects which is considerably less due to our experiences.

- *the set of task classes is decisive*

RKF can only be used in domains where a limited set of task classes can be found. If this set does not exist naturally for a domain (e.g. user groups) it might be a problem to find a set that is not too small (contradictory rewards would make learning impossible) and not too large (it becomes more difficult to classify a single task). In the worst case, each task class consists of only a single task which would also make training impossible.

- *dependencies are not learnt explicitly*

With RKF we only learn which objects to use in context of a task class *c*. Dependencies are only learnt in that way that the *objects* become relevant which fulfill these dependencies. The training processes of the objects involved in a dependency are more or less isolated. Especially dependencies that will affect the rewards of dependent objects are not allowed in RKF since RKF presumes constant or slow changing rewards for objects in context of a task class. For example dependencies like the following cannot be learnt: If A (90% reward) is selected then C will get a reward of 10% and if B (also 90% reward) is selected then C will get a reward of 70%. Both combinations could be learnt by RKF to be best, while a choice of B would result in a better sum of rewards.

- *user rewards*

RKF will only learn optimally if rewards for each component of the solutions are available. This might be a problem in complex systems where solutions consist of hundreds of components. Therefore we have made experiments with simplified assessment schemes. One is to rate the complete solution and use this value as the reward of all components of the solution. The simplest assessment scheme in this context can consist of just the values 0%= "system could **not** be sold" and 100%="system **has** been sold". With these simplified reward schemes the allowed combinations could be learnt but the training of a "good" solution is very slow and not possible in large domains.

# 5 Summary and Outlook

Our heuristic Relevant Knowledge First (RKF) has been developed based on a general study of potential improvements of knowledge-based search methods and was influenced by effects of cognitive psychology. First the method of selecting objects with RKF has been introduced, i.e. a random selection where the probability to select an object is proportional to its *relevance*. After that a function has been specified which calculates the relevance of objects. This function consists of two portions which describe the training and forgetting phases separately assuming that these phases occur alternately. Subsequently, the application of RKF in structure based configuration domains has been described and some results of our extensive experiments have been demonstrated. The specific characteristic of this heuristic is that objects can not only be *learnt*, but also be *forgotten* without being deleted. This can always be an advantage if a domain is subjected to changes which again is typical for most technical domains. Finally the effects arising from the use of RKF and limitations have been mentioned which have been observed during our experiments. In summary, it may be said that RKF could successfully be applied to all of our five different technical domains.

Future activities include the adaptation of the professional configuration system EngCon for our demands and the integration of RKF into it. EngCon was introduced in the "Workshop On Configuration" during the National Conference on Artificial Intelligence (AAAI) 1999 (see [Arlt et al.1999]).

## Biographical Notes

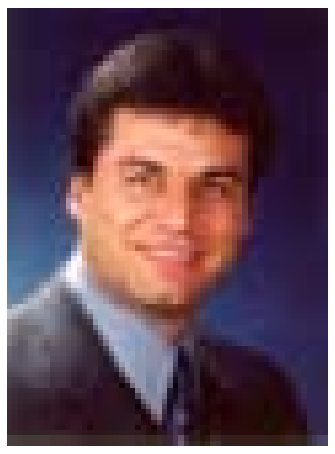

**Ingo Kreuz** studied Computer Science at the University of Stuttgart and graduated in 1996. In August 1997 he joined the Research and Technology department of DaimlerChrysler AG and works on his doctoral thesis about Knowledge Based Configuration. He is responsible for the ICON project (ICON = Intelligent Configuration) which is a co-operation between DaimlerChrysler AG and the University of Stuttgart.

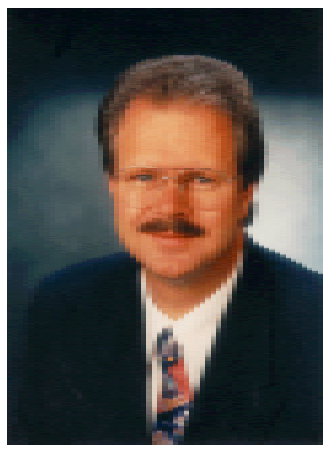

**Dieter Roller** is a member of the board of directors of the Institute of Computer Science at the University of Stuttgart. Further he is the head of the Graphical Engineering Systems Department and full professor of computer science. Additionally he has been awarded the distinction as a honorary professor of the University of Kaiserslautern.

Professor Roller holds an engineering diploma and received his PhD in computer science. As former research and development manager with world-wide responsibility for CAD-Technology within an international computer company, he gathered a comprehensive industrial experience. Professor Roller is well-known through many presentations throughout the world as well as through over 150 publications on computer-aided product development. Furthermore he has been granted several patents.

He is chairman of several national and international working groups and organiser of symposia, congresses and workshops. With his wealth of experience, he also serves as a technology consultant to various high-tech companies.